\documentclass{article}

\usepackage{microtype}
\usepackage{graphicx}
\usepackage{subfigure}
\usepackage{booktabs} 

\usepackage{amsmath,amsfonts,bm}

\def\eqref#1{(\ref{#1})}

\def\1{\bm{1}}

\DeclareMathAlphabet{\mathsfit}{\encodingdefault}{\sfdefault}{m}{sl}
\SetMathAlphabet{\mathsfit}{bold}{\encodingdefault}{\sfdefault}{bx}{n}

\def\gL{{\mathcal{L}}}

\newcommand{\E}{\mathbb{E}}

\usepackage{hyperref}
\usepackage{multirow}

\usepackage[linesnumbered,lined,boxed,commentsnumbered,ruled]{algorithm2e}

\usepackage[accepted]{icml2025}
\usepackage{diagbox}
\usepackage{amsmath}
\usepackage{amssymb}
\usepackage{mathtools}
\usepackage{amsthm}
\definecolor{red}{RGB}{247,77,77}
\definecolor{green}{RGB}{14,152,111}
\definecolor{purple}{RGB}{121,108,173}
\definecolor{orange}{RGB}{214,88,19}

\usepackage[capitalize,noabbrev]{cleveref}

\theoremstyle{plain}
\newtheorem{theorem}{Theorem}[section]

\theoremstyle{definition}
\newtheorem{definition}[theorem]{Definition}
\newtheorem{assumption}[theorem]{Assumption}
\theoremstyle{remark}
\newtheorem{remark}[theorem]{Remark}

\usepackage[textsize=tiny]{todonotes}

\begin{document}

\twocolumn[
\icmltitle{FedEGG: Federated Learning with Explicit Global Guidance}

\icmlsetsymbol{equal}{*}

\begin{icmlauthorlist}
\icmlauthor{Kun Zhai}{fudan}
\icmlauthor{Yifeng Gao}{fudan}
\icmlauthor{Difan Zou}{hongkong}
\icmlauthor{Guangnan Ye}{fudan}
\icmlauthor{Siheng Chen}{jiaoda}
\icmlauthor{Xingjun Ma}{fudan}
\icmlauthor{Yu-Gang Jiang}{fudan}
\end{icmlauthorlist}

\icmlaffiliation{fudan}{Fudan University, Shanghai, China}
\icmlaffiliation{jiaoda}{Shanghai Jiao Tong University, Shanghai, China}
\icmlaffiliation{hongkong}{The University of Hong Kong, Hong Kong, China}

\icmlcorrespondingauthor{Xingjun Ma}{xingjunma@fudan.edu.cn}

\icmlkeywords{Machine Learning, ICML}

\vskip 0.3in
]

\printAffiliationsAndNotice{\icmlEqualContribution}




\begin{abstract}
Federated Learning (FL) holds great potential for diverse applications owing to its privacy-preserving nature. However, its convergence is often challenged by non-IID data distributions, limiting its effectiveness in real-world deployments.
Existing methods help address these challenges via optimization-based client constraints, adaptive client selection, or the use of pre-trained models or synthetic data. In this work, we reinterpret these approaches as all introducing an \emph{implicit guiding task} to regularize and steer client learning. Following this insight, we propose to introduce an \emph{explicit global guiding task} into the current FL framework to improve convergence and performance.
To this end, we present \textbf{FedEGG}, a new FL algorithm that constructs a global guiding task using a well-defined, easy-to-converge learning task based on a public dataset and Large Language Models (LLMs). This approach effectively combines the strengths of federated (the original FL task) and centralized (the global guiding task) learning.
We provide a theoretical analysis of FedEGG’s convergence, examining the impact of data heterogeneity between the guiding and FL tasks and the guiding strength. Our analysis derives an upper bound for the optimal guiding strength, offering practical insights for implementation. Empirically, FedEGG demonstrates superior performance over state-of-the-art FL methods under both IID and non-IID settings, and further improves their performances when combined. 
\end{abstract}

\section{Introduction}\label{sec:Intro}

Federated Learning (FL) is a privacy-preserving training paradigm that enables multiple clients to collaboratively train a global model without sharing their private data \cite{kairouz2021advances}. Instead of exchanging raw data, FL shares gradients or model parameters, effectively addressing data privacy concerns. This approach has been successfully applied in various fields, including biometrics \cite{2021FedFace}, healthcare \cite{2020Federated}, and natural language processing \cite{2019Federated}. However, FL encounters significant challenges, such as high communication costs and slow model convergence, particularly when dealing with non-IID data. These challenges can lead to considerable gradient drifts, adversely affecting performance in real-world applications \cite{li2019convergence,zhao2018federated}.

To tackle FL convergence challenges under non-IID settings, several solutions have been proposed: (1) optimization-based client constraints, (2) adaptive client selection, and (3) leveraging pre-trained models or synthetic data. Empirical Risk Minimization (ERM) provides a framework for implementing client constraints by incorporating techniques like momentum and proximal terms to align local updates with global gradient directions. Notable methods include SCAFFOLD \cite{karimireddy2020scaffold}, which employs a control variable to mitigate client drift, and FedProx \cite{li2020federated}, which enhances local objective optimization by reducing divergence between local and global models. Additionally, \cite{cheng2023momentum} introduces momentum to improve the performance of FedAvg and SCAFFOLD, addressing the assumption of bounded data heterogeneity.
Another strategy is adaptive client selection. While random selection is straightforward, it may exacerbate data heterogeneity issues. Adaptive methods that consider factors such as client loss \cite{cho2020bandit, kim2020accurate} and local update volume \cite{ribero2020communication, chai2020tifl} can significantly reduce gradient aggregation variance. Furthermore, leveraging parameters from models pre-trained on large public datasets has been shown to enhance FL convergence. For instance, \cite{nguyen2022begin} highlights that pre-trained weights mitigate divergence on non-IID data, while FedPCL \cite{tan2022federated} provides a lightweight framework by integrating pre-trained models with prototype and contrastive learning. Additionally, synthetic data can also be utilized to reduce reliance on public datasets \cite{chen2022importance}. 

In this paper, we reinterpret the above improved FL methods as all introducing an \emph{implicit guiding task} into the FL process to regularize and steer client learning. Optimization-based client constraints leverage historical gradients to guide client updates, resembling a guiding task derived from past behaviors. Client selection methods explicitly construct a guiding task by prioritizing well-behaved clients, whose convergence leads others. Similarly, methods using pre-trained models or synthetic data introduce a strong initial bias, akin to starting from a pre-converged guiding task.
Building on this perspective, we hypothesize that explicitly introducing a server-side global guiding task can further enhance FL convergence and performance. If this global guiding task exhibits fast and stable convergence properties, it can improve learning dynamics—such as smoother optimization landscapes or richer representations—benefiting the overall FL process.

Motivated by this insight, we propose a novel FL algorithm named \textbf{FedEGG}, which explicitly introduces a global (server-side) guiding task into the existing FL framework. Specifically, it constructs the guiding task using a well-defined, easy-to-converge learning task based on a public dataset and Large Language Models (LLMs). Unlike the original FL task, the guiding task is designed solely to provide convergence guidance for the clients, forming a dual-task learning framework. In this setup, clients perform the original FL task while the server performs the centralized guiding task. By integrating the guiding task, FedEGG effectively combines the strengths of both federated and centralized learning. We provide a theoretical analysis of FedEGG’s convergence and empirically demonstrate its superiority over existing FL methods. 

In summary, our main contributions are:
\begin{itemize}
    \item  We propose \textbf{FedEGG}, a novel FL algorithm that constructs an explicit global guiding task with the help of LLMs to guide the convergence of the original FL task.

    \item We provide a theoretical analysis of FedEGG’s convergence advantage over FedAvg, deriving an upper bound for the optimal strength of the guiding task.

    \item  We empirically demonstrate FedEGG’s superior performance under non-IID settings, its ability to enhance existing FL methods when combined with them, and its improved convergence even under IID settings.
\end{itemize}


\section{Related Work}\label{sec:related}
We briefly review existing methods that address the convergence challenges of FL, particularly under non-IID data settings.

\noindent\textbf{Client Constraint Methods. }\; 
Convergence issues in FL arise from inconsistencies between local and global optima \cite{charles2021convergence, wang2021local, malinovskiy2020local}, making alignment between client and global objectives crucial. Momentum techniques, applied on the client side \cite{karimireddy2021breaking, xu2021fedcm}, server side \cite{wang2019slowmo, reddi2020adaptive}, or both \cite{khanduri2021stem, das2022faster}, accelerate convergence. For instance, FedCM \cite{xu2021fedcm} uses client-level momentum to address partial participation and heterogeneity, while SLOWMO \cite{wang2019slowmo} synchronizes clients and applies momentum updates iteratively. FedGLOMO \cite{das2022faster} achieves an \(\mathcal{O}(\epsilon^{-1.5})\) convergence rate by combining server and client momentum, reducing variance from client heterogeneity. Recent works like FedSGDA-M \cite{wu2024solving} and FEDAVG-M \cite{cheng2023momentum} further validate momentum’s benefits for FL convergence.  
Proximal term-based methods, such as FedProx \cite{li2020federated} and SCAFFOLD \cite{karimireddy2020scaffold}, control client updates to improve convergence. FedProx ensures local optimizations remain near the global model \cite{li2020federated}, with extensions showing success in autonomous driving \cite{donevski2021addressing} and computer vision \cite{he2021fedcv}. SCAFFOLD adjusts update directions based on client drift \cite{karimireddy2020scaffold}, preventing divergent local models from skewing the global model. FedDyn \cite{acar2021federated} introduces a dynamic regularizer to align local and global loss minima, reducing communication costs. FedFTG \cite{zhang2022fine} uses data-free knowledge distillation to enhance global models, while FedUR \cite{zhang2023fedur} optimizes updates using convergence upper bounds. Recently, FedCBO \cite{carrillo2024fedcbo} adopted consensus-based optimization with interacting particles, proving effective for non-convex objectives.

\noindent\textbf{Client Selection Methods. }\;  
Random client selection, though simple, often hinders FL due to data and device heterogeneity \cite{li2022pyramidfl}. Prioritizing "good" clients in each round \cite{lai2021oort} is more effective. Some methods focus on clients with higher local losses to accelerate convergence \cite{cho2020bandit, kim2020accurate}. For example, UCB-CS \cite{cho2020bandit} uses a bandit-based approach to optimize client selection, reducing communication costs while improving convergence. Similarly, FedCM \cite{kim2020accurate} employs a multi-armed bandit strategy for client sampling and combinatorial model averaging, addressing biased model averaging and improving generalization.  
Other approaches define client utility based on factors like model weights \cite{ribero2020communication} and training time \cite{chai2020tifl}. The method in \cite{ribero2020communication} uses an Ornstein-Uhlenbeck process to select clients with significant weight updates while estimating updates for non-selected clients, reducing communication overhead. TiFL \cite{chai2020tifl} organizes clients into tiers based on training performance, selecting clients from the same tier each round to mitigate the straggler problem. Its adaptive tier selection mechanism dynamically updates tiers, improving efficiency in heterogeneous environments.  
Oort \cite{lai2021oort}, a state-of-the-art client selection method, introduces a utility-based guiding scheme to address both data and system heterogeneity. Effective client selection must balance exploitation (prioritizing high performers) and exploration (diversifying selection) \cite{fu2023client}, aligning with principles seen in optimization-based methods.

\noindent\textbf{Pre-training-based Methods. }\;  
Pre-trained weights from large-scale public datasets \cite{hendrycks2019using, devlin2018bert} significantly improve model accuracy and robustness in centralized machine learning, a benefit that extends to FL \cite{nguyen2022begin}. Pre-trained weights transform FL into a fine-tuning process, mitigating divergence issues caused by non-IID data. For example, FedPCL \cite{tan2022federated} combines pre-trained models with prototype and contrastive learning to create a lightweight framework. Additionally, reliance on public datasets can be reduced by using synthetic data \cite{chen2022importance}. Pre-trained models provide a strong initialization for FL, reducing the gap between the initial and optimal models compared to random initialization, which is particularly beneficial given the divergence challenges posed by non-IID data.  
For instance, ViT-FL \cite{qu2022rethinking} integrates Vision Transformers into FedAvg, improving robustness on heterogeneous data. Prompt learning \cite{jia2022visual, zhou2022conditional, zhou2022learning} has also emerged as an efficient tuning method, enabling task-specific adaptation with minimal communication, making it well-suited for FL \cite{deng2024unlocking, bai2024diprompt, su2024federated}.

The above methods address FL convergence on non-IID data by either constraining local updates or leveraging pre-trained weights to stabilize convergence. In contrast, we explore a novel approach which introduces an \emph{explicit global guiding task} with the help of LLMs to regularize and steer client updates.

\section{Proposed Method}

\subsection{Problem Definition} 
We consider a typical FL setting \cite{mcmahan2017communication, mcmahan2016federated} with $N$ clients and a central server. Client $k \in [N]$ has its local dataset $D_k$. Note that $D_k$ may vary across different clients, reflecting client heterogeneity. 
Let $f_k$ be the local objective of the $k-$th client:  $f_k = \E_{\xi_k \sim D_k} [\gL(\mathbf{W}; \xi_k)]$. FL aims to learn a global model $\mathbf{W}$ over dataset $D=\cup_{k \in [N]} D_k$, via the following Empirical Risk Minimization (ERM):
\begin{equation}
    \min_{\mathbf{W}}   f(\mathbf{W}) = \textstyle\sum\nolimits_{k=1}^{N} p_k f_k(\mathbf{W}),
\end{equation}
where $p_k$ is the importance of the $k-$th client and $\sum_{k=1}^N p_k =1$. The notations used in this paper are detailed in Table \ref{tab:Summary-notation}.

\begin{table}[ht]
  \caption{Summary of notations used in this paper.}
  \label{tab:Summary-notation}
  \centering
  \resizebox{1\linewidth}{!}{ 
  \begin{tabular}{ll}
    \toprule
    \textbf{Notation}     &    \textbf{Description} \\
    \midrule
    \(N\), \(m\), \(k\)    &  \#clients, \#selected clients, the \(k\)-th client \\
    \(T\), \(t\) &   \#communication rounds, index of the current round \\
    \(T_c\), \(t_c\)  &   \#local update steps and step index \\
    \(T_g\), \(T_g\)  &    \#guiding task update steps and step index \\
    \(\overline{\mathbf{W}}_t\)    &  Aggregated model at round \(t\) \\
    \(\mathbf{W}_t\)        &   Global model at round \(t\) \\
    \(\mathbf{V}_{t}^k\)     &  Local model of the \(k\)-th client at round \(t\) \\
    \(\mathcal{L}(\mathbf{W}, D)\) &  Loss of model \(\mathbf{W}\) on dataset \(D\) \\
    \(\mathcal{L}^c\), \(\mathcal{L}^g\)  &  Loss of the FL task and guiding task, respectively \\
    \(\mathbf{g}_t^k\)   &   Mini-batch gradient of the \(k\)-th client at round \(t\) \\
    \(\boldsymbol{q}_t\)   & Mini-batch gradient of the guiding task at round \(t\) \\
    \(f\), \(f_k\), \(F\)    & Global, client \(k\), and guiding task functions \\
    \(\mathbf{W}^*\)   &  Optimal solution for the global \\ 
    \(\mathbf{W}_g^*\)   & Optimal solution for the guiding task \\
    \(\mathbf{V}_k^*\)   &  Optimal solution for the \(k\)-th client's task \\ 
    \(\eta_t\)  &     Learning rate for local client updates at round \(t\) \\
    \(\gamma_t\)   &  Learning rate for the guiding task at round \(t\) \\
    \(p_k\)     &      Importance weight of the \(k\)-th client \\
    \(\tau\)   &       Threshold for \(\log\left(\mathcal{L}^c/\mathcal{L}^g\right)\) \\
    \bottomrule
  \end{tabular}
  }
\end{table}

\subsection{FedEGG } 

\begin{figure*}[!ht]
  \centering
  \includegraphics[width=1\linewidth]{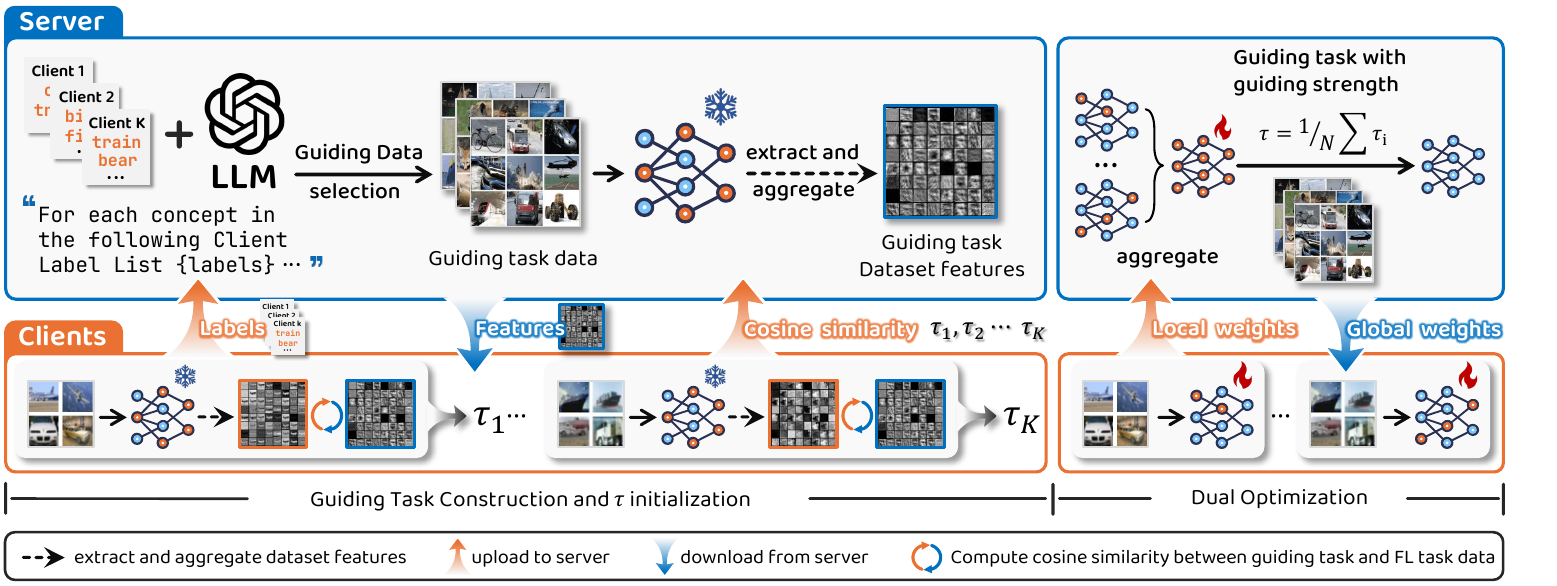}
   \vspace{-0.5cm}
  \caption{An overview of FedEGG. It operates in two phases: \emph{Phase 1:} Clients upload their data labels to the server, which are then used by a LLM to construct the guiding task. During each communication round, the server sends the averaged features of the guiding task to the clients, which are used to determine the guiding strength $\tau$.  \emph{Phase 2: } The FL task and guiding task are jointly optimized on the same model, with the guiding task providing explicit guidance to the FL task.}
  \label{Fig:pipeline}

\end{figure*}

\subsubsection{Overview. }
Figure \ref{Fig:pipeline} provides an overview of our proposed FedEGG method. It consists of two distinct phases: \textbf{Phase 1: Guiding Task Construction}, and \textbf{Phase 2: Dual Optimization}.

In \textbf{Phase 1}, the server constructs the guiding task by selecting an appropriate public dataset and initializing the guiding strength threshold $\tau$, with the assistance of a LLM. This phase ensures that the guiding task is well-defined and easy-to-converge, providing a robust foundation for guiding the FL process.
In \textbf{Phase 2}, the server collaborates with the clients to perform dual optimization. Here, the clients optimize the original FL task using their local data, while the server optimizes the guiding task in a centralized manner. The guiding task is specifically designed to provide convergence guidance to the clients, effectively steering their updates toward a more stable and efficient optimization trajectory. This dual-task framework allows FedEGG to combine the strengths of both federated and centralized learning, addressing the challenges of non-IID data and slow convergence in traditional FL settings.
Next, we detail each step of the FedEGG algorithm.

\begin{algorithm}[!htb]
   \SetAlgoLined
   \caption{FedEGG }
   \label{al:FedEGG}
   \KwIn{Initial global model $\mathbf{W}_0$; guiding task loss $\gL^g$ ;client learning rate $\eta_t$; guiding task learning rate $\gamma_t$; total client number $N$; total round number $T$; client step number $T_c$; guiding task step number $T_g$; client data $\{D_1,D_2,...,D_N\}$; guiding task data $D_g$; momentum $\beta$ }
   \KwOut{global model weight $\mathbf{W}_{T}$ }

    \underline{\textbf{Phase 1:}}\\
{\small (1): Clients in $[N]$ send labels to the server. \\
(2): The server selects guiding task data $D^g$ via GPT-4. \\
(3): The server computes and sends $F^l(\mathbf{W}_0, D^g)$  to clients. \\
(4): Each client $k \in [N]$ computes and sends $\tau_k$ to the server. \\
(5): The server computes $\tau$ (Eq. \eqref{eq:tau}).}

   \hspace{1cm}

\underline{\textbf{Phase 2:}}\\
\For{each round $t = 1,2,..., T$}{
     sample $m$ clients $C \subseteq \{1,2,...,N   \}$ \\
     broadcast $\mathbf{W}_{t-1}$ to all clients $k\in C$  \\
     
    \For{each client $k \in C$ }{
     optimizes the FL task: get $\mathbf{V}_k$ and $\gL_k$ (Eq. \eqref{eq:4}).
    }
    server aggregate: get $\overline{\mathbf{W}}_t$ (Eq. \eqref{eq:5}) and $\gL^c$ (Eq. \eqref{eq:6}). \\
    server optimize  guiding task: \\
    \If{$\log(\gL^c \big/ \gL^g) < \tau$}{
        \For{each step $t_g=1,2,...,T_g$}{
           update $\overline{\mathbf{W}}_t$ following (Eq. \eqref{eq:7})
        }
        $\gL^g = \gL(\overline{\mathbf{W}}_t,D^g) $ \\
    }
    $ \mathbf{W}_{t} = \overline{\mathbf{W}}_{t} $
}
\end{algorithm}

\subsubsection{Algorithm Description. } 
The detailed steps of FedEGG are summarized in Algorithm \ref{al:FedEGG}.
During \textbf{Phase 1}, it constructs the guiding task by selecting a subset of training data from a public dataset and  initializing the guiding strength threshold $\tau$. In this step, clients only need to upload the labels of their private data to the server. Using these labels, the server leverages a LLM to select the guiding task data $D^g$ from the public dataset. 
The server then uses the initial pre-trained global model $\mathbf{W}_0$ to extract features from the guiding task data $D^g$ and sends the averaged features from the $l$-th layer to the clients as $F^l(\mathbf{W}_0, D^g)$. This approach minimizes communication costs, as only averaged features are transmitted. Additionally, since the guiding task data originates from a public dataset, no privacy concerns arise from sharing these features.

Each client ($k$-th) uses the same pre-trained global model to extract features $f_k^l(\mathbf{W}_0, D_k)$ from its private data $D_k$ and computes the cosine similarity $\tau_k$ between $f_k^l(\mathbf{W}_0, D_k)$ and $F^l(\mathbf{W}_0, D^g)$. The similarity is calculated as:
\begin{equation}\label{eq-tau}
    \tau_k =\log\left[\cos \big(
    \frac{\textstyle\sum f_k^l(\mathbf{W}_0, D_k)}{|D_k|} , 
    F^l(\mathbf{W}_0, D^g) \big)\right].
\end{equation}
The guiding strength threshold $\tau$ is then computed based on $\tau_k$ for $k \in [N]$:
\begin{equation} \label{eq:tau}
    \tau = \rho \cdot \frac{1}{N} \textstyle \sum_k \tau_k + \iota ,
\end{equation}
where $\rho$ and $\iota$ are temperature hyper-parameters. A higher similarity between the FL task and the guiding task data results in a larger $\tau$. As discussed in Section \ref{sec:convergence}, the key factor determining $\tau$ is the data heterogeneity between the guiding task and the FL task, denoted as $\Gamma_g$.

During \textbf{Phase 2}, the server manages three types of information: 1) the global model $\mathbf{W}_t$, 2) the guiding task data $D^g$ selected from the public dataset, and 3) the training loss of the client models $\gL^c$. In the $t$-th communication round, the server broadcasts the global model $\mathbf{W}_t$ to the selected clients $C \subseteq [N]$. Each client $k \in C$ initializes its local model with the received global model, setting $\mathbf{V}_k = \mathbf{W}_t$, and performs local updates as:
\begin{equation}\label{eq:4}
    \mathbf{V}_k = \mathbf{V}_k -\eta_t \nabla f_k(\mathbf{V}_k,\xi_k), \;\; \gL_k = \gL(\mathbf{V}_k,D_k),
\end{equation}
where $\xi_k$ is a sample uniformly chosen from the local dataset $D_k$, and $\gL_k$ is the training loss of the $k$-th client.

After a predefined number of local updates, the clients upload their accumulated updates to the server. The server aggregates the local updates and calculates the momentum of local losses as:
\begin{equation}\label{eq:5}
    \overline{\mathbf{W}}_t = (1 \big/ m) \textstyle\sum\nolimits_{k\in C}\mathbf{V}_k,
\end{equation}
\begin{equation}\label{eq:6}
    \gL^c = \beta *\gL^c + (1-\beta) (1 \big/ m) \textstyle\sum\nolimits_{k\in C} \gL_k,
\end{equation}
where the momentum technique is employed to accelerate training.

Next, the server optimizes the guiding task by updating the global model as:
\begin{equation}\label{eq:7}
    \overline{\mathbf{W}}_t = \overline{\mathbf{W}}_t - \gamma_t \nabla F(\overline{\mathbf{W}}_t,\xi^g).
\end{equation}
To ensure that the guiding task does not conflict with the FL task, we introduce the \emph{Log Loss Ratio} (LLR) constraint:
\begin{equation}\label{eq:loss-ratio-constraint}
    \log\big(\gL^c \big/ \gL^g\big) < \tau,
\end{equation}
where $\gL^g$ is the loss of the guiding task and $\tau$ is the guiding strength threshold. This constraint ensures a balanced influence of the guiding task on the FL process.
Finally, the server updates the global model for the next communication round as $\mathbf{W}_{t+1} =\overline{\mathbf{W}}_t$.

\subsubsection{Guiding Task. }
Selecting an appropriate dataset for the guiding task is critical to the success of FedEGG. A smaller $\Gamma_g$ enhances the guiding effect; in the extreme case where the guiding task and the FL task are identical, FL effectively reduces to traditional centralized machine learning, where the server has access to the clients' full training data. More generally, the guiding task should utilize a public dataset that closely resembles the FL task. Intuitively, smaller $\Gamma_g$ values lead to better convergence. In FedEGG, the selection process only requires analyzing task information (e.g., class names) of the FL task, which can be easily automated using a LLM. For example, in an FL task based on the CIFAR-10 dataset, a subset of classes (e.g., 20) from the ImageNet-1k \cite{deng2009imagenet} dataset could be selected to construct the guiding task. Our experiments evaluate three different selection strategies using GPT-4 \cite{achiam2023gpt}, focusing on the degree of concept overlap between the guiding task and the FL task, as detailed in Section \ref{sec:exp setup}. While some FL tasks may easily identify a guiding task with low heterogeneity, others may face challenges. This can be partially addressed by leveraging text-to-image generative models or multimodal large language models (MLLMs) such as Stable Diffusion 3.0 \cite{esser2024scaling} and GPT-4 as described in Section \ref{sec:main result}.

\subsubsection{Guiding Strength. }
The log loss ratio (LLR) $\log(\gL^c/\gL^g)$ in Eq. \eqref{eq:loss-ratio-constraint} controls the guiding strength. A higher LLR corresponds to a lower loss ($\gL^g$) for the guiding task, indicating stronger guidance. While the guiding task is designed to be well-defined and converge efficiently, its primary role is to enhance convergence by creating a smoother loss landscape, rather than directly aligning with the FL task. However, an excessively high guiding strength (i.e., a higher LLR) may bias the global model toward the guiding task, potentially hindering the convergence of the original FL task. Therefore, it is essential to carefully select the threshold $\tau$ to balance accelerated convergence and improved generalization. As data heterogeneity $\Gamma_g$ increases, the guiding strength should be weakened. In Eq. \eqref{eq-tau}, a larger $\Gamma_g$ reduces the cosine similarity between the data features of the guiding task and the FL task, thereby decreasing $\tau$.

\subsection{Convergence Analysis}\label{sec:convergence}
We begin by outlining the assumptions and definitions in line with previous work \cite{li2019convergence}, followed by a convergence analysis of FedEGG.

\begin{assumption}\label{assump:1}
The functions $f_1,...,f_N$and $F$ are all $L$-smooth: for all $V$ and $W$, $f_k(V) \leq f_k(W) + (V-W)^{\textsuperscript{T}} \nabla f_k(W) + \frac{L}{2}||V-W||_2^2$
\end{assumption}

\begin{assumption}\label{assump:2}
The functions $f_1,...,f_N$ and $F$ are all $\mu$-strongly convex: for all $V$ and $W$, $f_k(V) \geq f_k(W) + (V-W)^{\textsuperscript{T}} \nabla f_k(W) + \frac{\mu}{2}||V-W||_2^2$
\end{assumption}

\begin{assumption}\label{assump:3}
Let $\xi_t^k$ be sampled uniformly at random from the $k$-th client's local data. The variance of stochastic gradients for each client is bounded: $\mathbb{E} \big[\| \nabla f_k(\mathbf{V}_t^k, \xi_t^k)- \nabla f_k(\mathbf{V}_t^k) \|^2 \big] \leq \sigma_k^2$ for $k=1, 2, ..., N.$
\end{assumption}

\begin{assumption}\label{assump:4}
Let $\xi_t^g$ be sampled uniformly at random from the server data. The variance of stochastic gradients is bounded: 
$\mathbb{E} \big[\| \nabla F(\overline{\mathbf{W}}_t, \xi_t^g)- \nabla F(\overline{\mathbf{W}}_t) \|^2 \big] \leq \sigma_g^2 $
\end{assumption}


\begin{definition}\label{definition:2}
[FL-Guiding Task Heterogeneity] Let $f^*$ and $F^*$ be the optima of the FL $f$ task and the guiding task $F$, respectively. The heterogeneity between the two tasks are defined as $\Gamma_g = f^* - F^*$. If the data is IID between the two tasks, then $\Gamma_g = 0 $. If the data is non-IID, $\Gamma_g$ is nonzero, and its magnitude measures the heterogeneity between the two tasks.
\end{definition}

\begin{definition}\label{definition:3}
[Guiding Strength] Let $\overline{\mathbf{W}}_t$ be the aggregated model in the $t-$th communication round, the guiding strength of the guiding task $F$ on the FL task $f$ is defined as 
$\Pi = f^* - F(\overline{\mathbf{W}}_t)$, where $F(\overline{\mathbf{W}}_t)$ is the performance of the aggregated model on the guiding task, and $f^*$ is the optimal solution of the FL task. As $\Pi$ increases, it indicates a stronger guiding strength. 
\end{definition}

\noindent \textbf{Note:} To prevent ``overguiding" caused by an excessively strong guiding task, we design the guiding task to satisfy \( f^* \ll F(\overline{\mathbf{W}}_t) \), ensuring that \( \Pi < 0 \). This implies that the guiding task has a higher training error compared to the original FL task.

\subsubsection{Convergence Result. }
We focus on the case of full participation where all clients participate in the global aggregation. Following \cite{li2019convergence}, we define
\begin{align*}
\Delta_{t+1}^{FedEGG} &= \mathbb{E}\big[  \|\mathbf{W}_{t} -  \eta_t\mathbf{g}_t -\gamma_t \boldsymbol{q}_t -\mathbf{W}^* \|^2 \big], \\
\Delta_{t+1}^{FedAvg} &= \mathbb{E}\big[  \|\mathbf{W}_{t}  - \eta_t \mathbf{g}_{t} - \mathbf{W}^* \|^2 \big],
\end{align*}
where \(\mathbf{W}_{t+1} = \mathbf{W}_t - \eta_t\mathbf{g}_t -\gamma_t \boldsymbol{q}_t\) denotes one-step update of FedEGG, and \(\mathbf{W}_{t+1} = \mathbf{W}_t  -\eta_t\mathbf{g}_t\) denotes the corresponding update of FedAvg. The additional term \(\gamma_t \boldsymbol{q}_t\) in FedEGG represents the update from the guiding task, enabling FedEGG to guide FedAvg toward a better optimal solution.

\begin{theorem}\label{them:FedEGG}
Assume Assumptions 1-4 hold and let \(\eta_t \leq \frac{1}{4L}\). Then, we have:
\begin{flalign*}
   \Delta_{t+1}^{FedEGG} = \Delta_{t+1}^{FedAvg} + \epsilon,
\end{flalign*}
where $\epsilon = \gamma_t^2\sigma_g^2  + 2\left(\frac{1}{\mu} - 2\gamma_t(1 - L\gamma_t)\right) \Gamma_g  + 4\gamma_t(1 - L\gamma_t) \Pi,$
with \(\Gamma_g = f^* - F^*\) and \(\Pi = f^* - F(\overline{\mathbf{W}}_t)\).
\end{theorem}

The proof is provided in Appendix \ref{sec:A}. Theorem \ref{them:FedEGG} shows that FedEGG introduces an additional term \(\epsilon\) compared to FedAvg, reflecting the impact of the guiding task on its convergence. Here, \(\mu\), \(L\), and \(\Gamma_g\) are constants, and \(\sigma_g^2\) represents the variance of the stochastic gradient, making \(\Pi\) a crucial factor. During the guiding phase, with \(f^*\) fixed, the model \(\overline{\mathbf{W}}_t\) is trained on the FL task. When the guiding task is active, \(F(\overline{\mathbf{W}}_t)\) becomes significantly large (i.e., \(f^* \ll F(\overline{\mathbf{W}}_t)\)), resulting in \(\epsilon < 0\) and thereby improving FedEGG's convergence. As the guiding task progresses, \(F(\overline{\mathbf{W}}_t)\) decreases and \(\Pi\) increases until \(\epsilon\) reaches zero, indicating that the guiding task no longer contributes to convergence and should be terminated. This raises an important question: \textit{How can we effectively determine when to stop the guiding task?}

\subsubsection{Remarks. }

\begin{remark}\label{mark1}
[Relationship between $\Pi$ and $\Gamma_g$] During the guiding phase in theorem \ref{them:FedEGG}, $\epsilon <0 $ , which implies:
\begin{equation} \label{Eq8}
    \Pi \leq   -\big ( \frac{1}{2\mu \gamma_t(1 - L\gamma_t)} -  1 \big)\Gamma_g -\frac{\gamma_t^2\sigma_g^2 }{4\gamma_t(1 - L\gamma_t)},
\end{equation}
where $\gamma_t < \frac{1}{L}$ and $2\mu \gamma_t(1 - L\gamma_t) > 1$.
\end{remark}

Eq. \eqref{Eq8} indicates that a larger $\Gamma_g$ tends to leads to a smaller upper bound for $\Pi$. I.e., higher FL-Guiding task heterogeneity necessitates an earlier termination of the guiding task, and vice versa. This supports the rationale behind our Eq. \eqref{eq-tau}, which calculates $\tau$ (the threshold for terminating the guiding task) as follows: 
\noindent {\small $  \tau =\log\Big [\rho \cdot \cos \big( \frac{1}{N} \sum_{k\in [N]} 
    \frac{\textstyle\sum f_k^l(\mathbf{W}_0, D_k)}{|D_k|} , F^l(\mathbf{W}_0, D^g) \big) + \iota \Big]$.} 
In other words,  $\tau$ is determined by the cosine similarity between the guiding task data ($D^g$) and the FL task data ($D_k$). When data heterogeneity is high, this similarity decreases, leading to a smaller $\tau$ and consequently prompting an earlier termination of the guiding task. 

\noindent\textbf{Note:}  In the ideal case, when $\Gamma_g= 0$, we have $\epsilon = \gamma_t^2\sigma_g^2  + 4\gamma_t(1 - L\gamma_t) \Pi$.
Under this condition, $f$ equals $F$, leading to $\Pi = f^* - F(\overline{\mathbf{W}}_t) = f^* - f(\overline{\mathbf{W}}_t) \leq 0$. Moreover, as $\sigma_g$ approaches 0, $\epsilon <0$ consistently holds. This indicates that when the data is homogeneous, the influence of the guiding task on FedEGG is always beneficial, which aligns with the effectiveness of centralized learning.

Finally, to better understand the relationship between the guiding task and FedEGG's convergence, consider a scenario where the guiding task provides no guidance to the FL task. This leads to the following remark:

\begin{remark}\label{mark2}
[Consistency] If the guiding task does not provide guidance to the FL task, it follows that $\epsilon =0$, where $\gamma_t = 0 $, $\Gamma_g =0$, $\Pi =0$ and $\sigma_g^2 =0$. In this case, we have:
\begin{flalign*}
   \Delta_{t+1}^{FedEGG} = \Delta_{t+1}^{FedAvg}.
\end{flalign*}
\end{remark}

When the guiding task does not guide the FL task, all related parameters  ($\gamma_t$, $\Gamma_g$, $\Pi$ and $\sigma_g^2$) are zero. Consequently, the convergence of FedEGG aligns with that of the standard FedAvg. This consistency ensures that, in the absence of additional tasks, FedEGG does not introduce any extra bias, maintaining its alignment with the traditional method.

\section{Experiments}\label{sec: experiments}

\subsection{Experimental Setup} \label{sec:exp setup}

\noindent{\textbf{Datasets and Models.}} \;
We consider two widely adopted benchmark datasets for FL: CIFAR-10 and CIFAR-100 \cite{krizhevsky2009learning} with heterogeneous dataset partitioning. Following prior work \cite{li2022federated}, we employ the Dirichlet distribution \textbf{Dir($\alpha$)} to simulate non-IID settings, where smaller values of $\alpha$ indicate higher data heterogeneity. In our main experiments, we set $\alpha=0.1$ (non-IID) and $\alpha=10$ (IID) for CIFAR-10, and $\alpha=0.01$ (non-IID) and $\alpha=100$ (IID) for CIFAR-100. For both datasets, we initialize all methods using ResNet18 \cite{he2016deep} with pre-trained weights on ImageNet-1k.

\noindent{\textbf{Baselines.}} \;
We compare against 7 FL baselines: 1) \emph{FedAvg} \cite{mcmahan2017communication}, a classic FL algorithm; 2) \emph{SCAFFOLD} \cite{karimireddy2020scaffold}, which reduces client drift by incorporating control variates into local updates; 3) \emph{FedProx} \cite{li2020federated}, which enhances local optimization by regularizing the divergence between local and global models; and 4) \emph{FedNova} \cite{wang2020federated}, which normalizes and scales local updates to address object inconsistency while ensuring convergence; 5) \emph{FedDyn} \cite{acar2021federated}, which introduces a dynamic regularizer to align global and local solutions while reducing communication costs; 6) \emph{FedDC} \cite{gao2022feddc}, which uses
control variates to correct each client gradients but incurs
double communication costs; and 7) \emph{FedFTG} \cite{zhang2022fine}, which refines the global model through an additional fine-tuning step using hard examples.

\noindent{\textbf{FedEGG Implementation.}} \;
We define the guiding task using ImageNet-1k\cite{deng2009imagenet}, making it applicable to a wide range of vision tasks. Specifically, we explore three types of guiding tasks: \textbf{High Heterogeneity (HH)}, \textbf{Medium Heterogeneity (MH)}, and \textbf{Low Heterogeneity (LH)}. For CIFAR-10, the LH guiding task is constructed by identifying the top 20 ImageNet classes with concepts most similar to those in the FL task, as determined by GPT-4. For the MH guiding task, we replace the top 10-20 classes with 10 randomly selected classes from ImageNet-1k. For the HH guiding task, we randomly select 20 classes from ImageNet-1k. For CIFAR-100, we follow a similar procedure to construct LH, MH, and HH guiding tasks, each comprising 40 classes. \emph{Note that the entire construction process relies solely on class names.}.

\noindent{\textbf{Hyper-parameters.}} \;
For all FL methods, we set the total number of clients to \(N = 100\), the number of sampled clients per round to \(m = 20\), and the total communication rounds to \(T = 300\). For the FL tasks, we set the local steps to \(T_c = 5\) with a batch size of 32. The learning rate \(\eta_t\) is set to \([1e-2, 1e-3, 1e-4]\) for \(t < 100\) and \(1e-5\) otherwise. The base of \(\log(.)\) in LLR is set to 2. In each experiment, we evaluate the average performance over the last 50 rounds. For all guiding tasks, we adopt local steps \(T_g = 1\), a batch size of 64, and a learning rate \(\gamma_t = \eta_t\). When calculating \(\tau\) in Eq. (\ref{eq-tau}), we use features extracted from the last convolutional layer to compute the cosine similarity, with \(\iota = -0.5\) and \(\rho = 2\).

\noindent{\textbf{Guiding Task Construction.}}  \;
We assume the server has access to meta-information about the FL task, such as the class names of the classification task. If class names are unavailable, the server can request clients to upload their associated concepts. Using this meta-information, the server can identify the most similar concepts in a public dataset (e.g., ImageNet-1k) with the assistance of GPT-4. In scenarios where no closely related public datasets are available, the server can utilize a diffusion model to generate images representing similar concepts. However, in this work, we focus exclusively on the case where a large-scale public dataset is available to construct the guiding task. The detailed prompts and returned class names by GPT-4 can be found in Appendix \ref{sec:B}.

\subsection{Main Results} \label{sec:main result}

\noindent{\textbf{Effectiveness of FedEGG. }}\;
Table \ref{tab:Average acc} reports the average training and test accuracies on the CIFAR-10 and CIFAR-100 datasets under both non-IID and IID settings. FedEGG consistently outperforms all baseline methods across all scenarios. Under the non-IID setting, FedEGG achieves significant improvements, increasing test accuracy by \textbf{3.39\%} on CIFAR-10 and \textbf{1.23\%} on CIFAR-100, compared to the second-best method. Although the performance gains are modest under the IID setting, FedEGG still demonstrates a clear advantage. Notably, FedFTG, which relies on generated virtual data (hard samples) to fine-tune the global model, performs poorly, particularly when applied to pre-trained models. This highlights the effectiveness of our FedEGG in leveraging explicit global guidance for improved convergence and accuracy.

\begin{table}[!htb]
  \caption{Average training and test accuracies (\%) of different methods on CIFAR-10 and CIFAR-100 under both non-IID and IID settings. \textbf{Bold} denotes the best results, \underline{ } indicates the second-best results, and +/- represents the performance increase/decrease between FedEGG and the best baseline. }
  \label{tab:Average acc}
  \centering
  \resizebox{1\linewidth}{!}{ 
  \begin{tabular}{cc|cc|cc|cc|ccl}
    \toprule
    \multicolumn{2}{c|}{\multirow{3}{*}{\textbf{Method}}}    &   \multicolumn{4}{c|}{\textbf{CIFAR-10}}  &    \multicolumn{4}{c}{\textbf{CIFAR-100}}  \\
             &    &   \multicolumn{2}{c|}{non-IID (Dir(0.1))}   &  \multicolumn{2}{c|}{IID }  
               &   \multicolumn{2}{c|}{non-IID (Dir(0.01))}   &  \multicolumn{2}{c}{IID }      \\
            &  &   Training  & Test &   Training  & Test &   Training  & Test &   Training  & Test \\
               
    \midrule
    \multicolumn{2}{c|}{FedAvg}     &   \underline{92.66}    &    \underline{88.06}    &    99.99     &    \underline{95.58}   &   67.13    &    58.53    &    99.97     &    80.03  \\
   \multicolumn{2}{c|}{SCAFFOLD}    &   84.92  &  82.84  &  \underline{100}   &  94.99   &  62.17   &    52.66    &    99.12     &    78.59   \\
    \multicolumn{2}{c|}{FedProx}    &   90.97  &  85.23  &  99.99   &    95.48  &    67.18 &   58.72    &    \underline{99.98}     &    \underline{80.13}   \\
    \multicolumn{2}{c|}{FedNova}   &   91.70  &  86.32  &  99.99   &    95.46   &   \underline{69.93}  &   \underline{60.12}   &    99.98     &    80.11      \\
    \multicolumn{2}{c|}{FedDyn}   & 83.11     &   80.43     &    99.90     &     93.02     &    46.30     &    42.87  &    99.59       &   72.61     \\
    \multicolumn{2}{c|}{FedDC}    & 88.48        &   81.52     &    99.99     &     88.98     &    52.92     &    34.89  &    99.98       &   65.76     \\
     \multicolumn{2}{c|}{FedFTG}    &  58.06        &   54.87     &    97.44     &     78.879     &    53.03     &    37.53  &        91.09  &   56.42     \\
    \midrule
    \multicolumn{2}{c|}{\textbf{FedEGG}}    &   \textbf{95.26}  &  \textbf{91.45} &  \textbf{100}   &    \textbf{95.88} &  \textbf{71.22}     &  \textbf{61.35}    &    \textbf{99.98}    &    \textbf{80.38}  \\
    & & (+2.6)  & (+3.39)  & (+0)   & (+0.03) & (+1.29)  & (+1.23)  & (+0) & (+0.25) \\   
     \bottomrule    
  \end{tabular}
}
\end{table}

\begin{table}[!htb]
  \caption{Average test accuracies (\%) of baseline FL methods \emph{combined with our FedEGG} on CIFAR-10 and CIFAR-100 datasets under both non-IID and IID settings. The guiding task levels are LH, MH, and HH. The +/- values indicate the accuracy increase or decrease when comparing FedEGG with FedAvg (with/without FedEGG).}
  \label{tab:Acc_Combined}
  \centering
  \resizebox{1\linewidth}{!}{ 
  \begin{tabular}{c|cc|c|c|c|c}
    \toprule
    & \multicolumn{2}{c|}{\multirow{3}{*}{\textbf{Method}}}    &   \multicolumn{2}{c|}{\textbf{CIFAR-10}}  &    \multicolumn{2}{c}{\textbf{CIFAR-100}}  \\
    &  &    &   non-IID (Dir(0.1)   &  IID   
    &    non-IID (Dir(0.01))   &  IID      \\
    \midrule
  \multirow{6}{*}{\rotatebox{90}{\textbf{LH}}}  &  \multicolumn{2}{c|}{\textbf{FedEGG}}    &    91.45 (+3.39)  &    95.88 (+0.3)     &  61.35  (+2.82)      &    80.38 (+0.35) \\
   &  \multicolumn{2}{c|}{SCAFFOLD + \textbf{FedEGG}}    &  86.48 (+3.64)   &    96.58 (+1.59)    &    57.04 (+4.38)    &    82.46 (+3.87)      \\
   &    \multicolumn{2}{c|}{FedProx +  \textbf{FedEGG}}  &  91.41 (+6.18)    &    95.82 (+0.34)      &    61.65 (+2.92)      &    80.84 (+0.71)      \\
   &  \multicolumn{2}{c|}{FedNova +  \textbf{FedEGG}}  &  91.89 (+5.57)   &    95.82 (+0.36)       &    63.27 (+3.15)    &    80.84 (+0.73)      \\
   &   \multicolumn{2}{c|}{FedDyn +  \textbf{FedEGG}}     &   84.05 (+3.61)    &     94.42 (+1.4)     &    47.56 (+4.69)        &   74.66 (+2.05)     \\
   &   \multicolumn{2}{c|}{FedDC +  \textbf{FedEGG}}      &   82.39 (+0.87)     &    90.14 (+1.16 )   &       35.50 (+0.61)  &     66.74  ( +0.98 )\\
    \midrule
  \multirow{6}{*}{\rotatebox{90}{\textbf{MH}}} & \multicolumn{2}{c|}{ \textbf{FedEGG}}    &    90.71 (+2.65)  &     95.78 (+0.02)    &     59.24 (+0.71)   &     80.13 (+0.10)      \\
   &  \multicolumn{2}{c|}{SCAFFOLD + \textbf{FedEGG}}    &  83.73 (+0.88)  &   95.52 (+0.53)    &      55.69 (+3.03)  &      80.15 (+1.56)      \\
    &   \multicolumn{2}{c|}{FedProx +  \textbf{FedEGG}}  &   87.15 (+1.92)   &   95.58 (+0.10)   &       58.92 (+0.20)   &      80.26 (+0.13)      \\
   &  \multicolumn{2}{c|}{FedNova +  \textbf{FedEGG}}   &   87.88 (+1.56)  &    94.87 (-0.39)    &    60.25 (+0.13)   &     79.81 (-0.29)      \\
    &  \multicolumn{2}{c|}{FedDyn +  \textbf{FedEGG}}  &    81.27 (+0.84)  &  94.05 (+1.03)   &      44.07 (+1.20)   &     75.29 (+2.68)     \\
   &   \multicolumn{2}{c|}{FedDC +  \textbf{FedEGG}}    &  81.98 (+0.46)  &    90.98 (+2.0)    &     38.72 (+3.83)   &    68.28 (+2.52)      \\
    \midrule
  \multirow{6}{*}{\rotatebox{90}{\textbf{HH}}} & \multicolumn{2}{c|}{ \textbf{FedEGG}}    &   89.9 (+1.84) &     95.73 (+0.15)   &       57.54 (-0.99)   &       78.8 (-1.23)      \\
   &  \multicolumn{2}{c|}{SCAFFOLD + \textbf{FedEGG}}     &    82.98 (+0.15)   &   95.23 (+0.24)    &     52.92 (+0.26)   &     79.99 (+1.39)      \\
    &   \multicolumn{2}{c|}{FedProx +  \textbf{FedEGG}}  &  86.98 (+1.75)    &   94.57 (-0.91)    &   58.03  (-0.67)   &     78.87 (-1.25)      \\
   &  \multicolumn{2}{c|}{FedNova +  \textbf{FedEGG}}   &    87.01 (+0.69)  &  94.60 (-0.85)    &      58.34 (-1.77)  &      78.85 (-1.26)      \\
   &   \multicolumn{2}{c|}{FedDyn +  \textbf{FedEGG}}   &   80.81 (+0.38)  &   93.56 (+0.54)    &      44.27 (+1.4)  &     74.97 (+2.36)      \\
   &   \multicolumn{2}{c|}{FedDC +  \textbf{FedEGG}}   &    83.70 (+2.18)   &   90.84 (+1.85)   &      40.0 (+5.11)   &      68.32 (+2.56)      \\
    \bottomrule    
  \end{tabular}
}
\end{table}

\noindent{\textbf{Effectiveness When Combined With Existing Methods. }}\;
We evaluate whether FedEGG can be combined with existing FL methods to further enhance their performance. Table \ref{tab:Acc_Combined} presents the results for guiding task levels ranging from LH to MH and HH. 
In the LH case, combining FedEGG with baselines significantly improves their performance. On CIFAR-10 under the non-IID setting, FedEGG increases test accuracy for SCAFFOLD, FedProx, FedNova, FedDyn, and FedDC by \textbf{3.64\%}, \textbf{6.18\%}, \textbf{5.57\%}, \textbf{3.61\%}, and \textbf{0.87\%}, respectively. On CIFAR-100, the improvements are \textbf{4.38\%}, \textbf{2.92\%}, \textbf{3.15\%}, \textbf{4.69\%}, and \textbf{0.61\%}, respectively. Under the IID setting, FedEGG boosts SCAFFOLD's test accuracy by \textbf{1.59\%} on CIFAR-10 and \textbf{3.87\%} on CIFAR-100, with smaller but consistent gains for other methods.
Similar trends are observed in the MH and HH cases, though the improvements are less pronounced compared to the LH case. In some instances, FedEGG fails to enhance performance, likely due to stability issues of the algorithm under high-heterogeneity guidance. 
Overall, lower task heterogeneity leads to greater improvements in baseline convergence with FedEGG, particularly in non-IID settings. These findings align with our theoretical analysis, highlighting the importance of constructing appropriate guiding tasks.

\begin{figure}[H]
  \centering
  \vspace{-0.1cm}
  \includegraphics[width=0.95\linewidth]{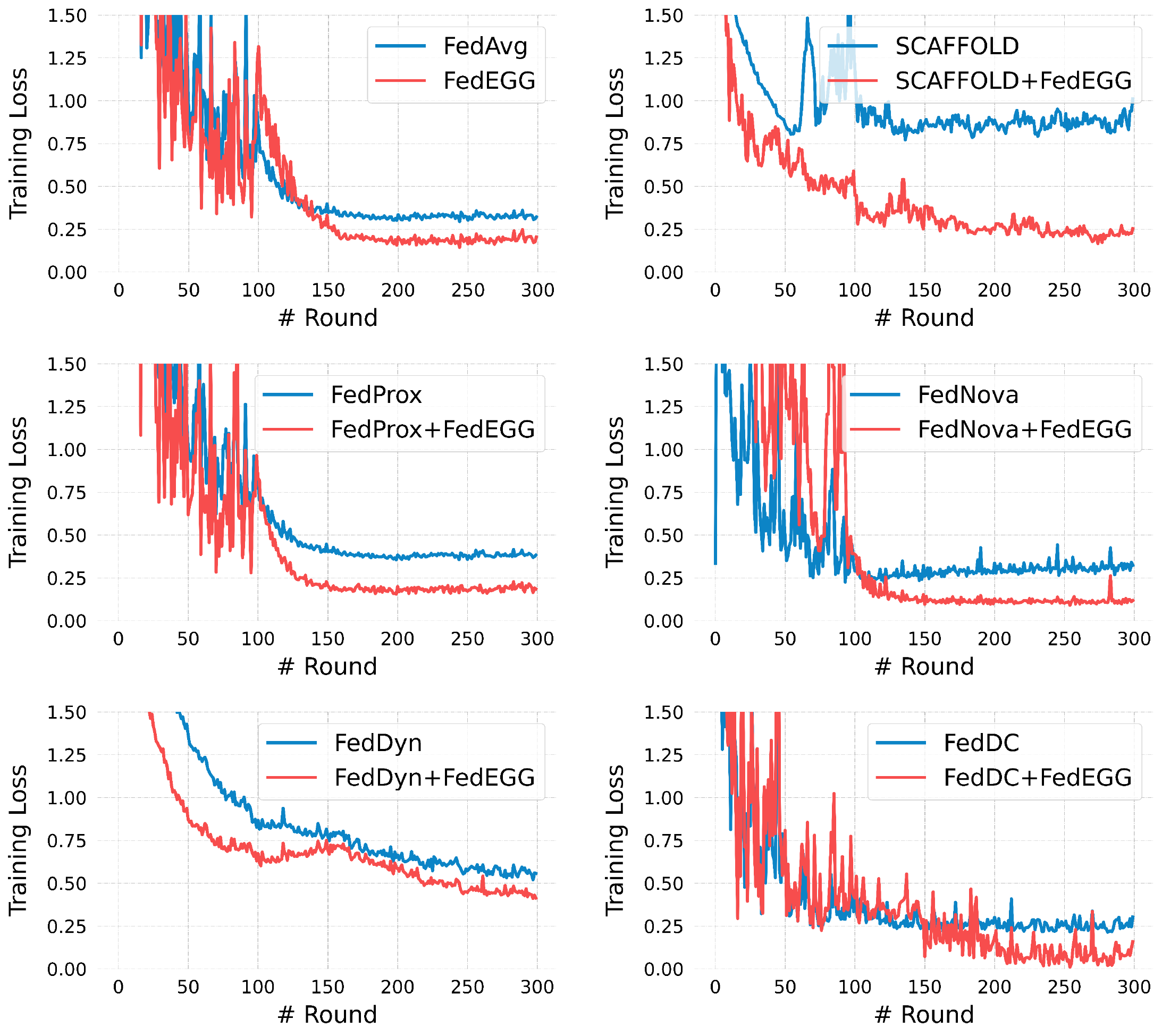}
  \vspace{-0.3cm}
  \caption{ The training losses of different FL methods across 300 rounds on CIFAR-10 under the non-IID setting.}
  \label{Fig:loss_noniid}
  \vspace{-0.2cm}
\end{figure}

\noindent{\textbf{Convergence Acceleration. }}\; Figure \ref{Fig:loss_noniid} demonstrates the accelerating effect of FedEGG on the convergence of baseline methods for CIFAR-10 under the non-IID setting. During the guiding phase, the convergence speed of the FL task initially slows. However, once the guiding task is completed, the FL task rapidly converges within a few training iterations. This highlights the importance of the guiding task converging first to effectively assist the FL task. 
Additional convergence plots for IID CIFAR-10, non-IID CIFAR-100, and IID CIFAR-100 are provided in the Appendix \ref{sec:C}. These plots consistently show that combining baseline methods with FedEGG reduces training loss. Notably, the sharp drops in training loss correspond to reductions in the learning rate by a factor of 10.

\begin{figure}[H]
  \centering
  \includegraphics[width=0.65\linewidth]{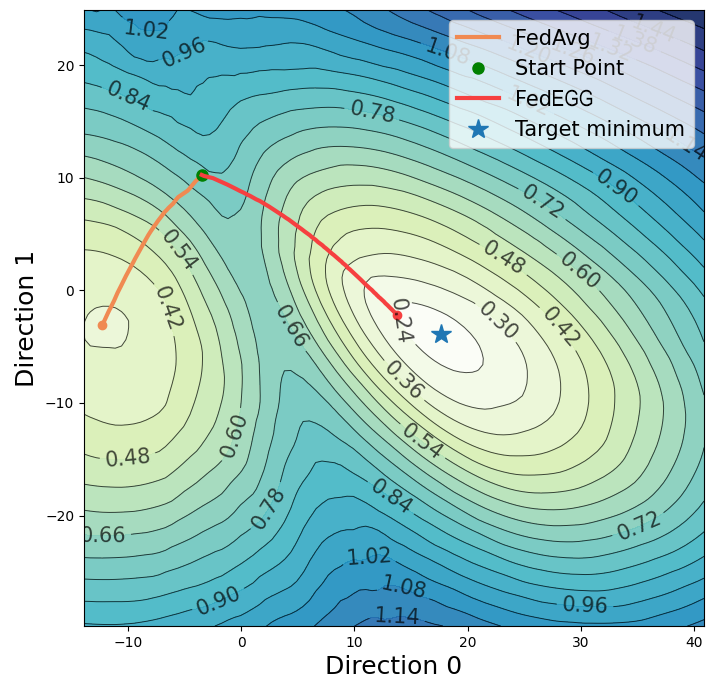}
  \vspace{-0.4cm}
  \caption{ Optimization trajectories of FedAvg and FedEGG in the loss landscape on CIFAR-10 (non-IID). }
  \label{Fig:lanscape}
  \vspace{-0.1cm}
\end{figure}

\noindent{\textbf{Loss Landscape. }}\;
In Figure \ref{Fig:lanscape}, we further show that the existence of the guiding task can help smooth the loss landscape of the FL task. Here, we visualize the optimization trajectory of FedAvg and FedEGG, with FedEGG demonstrating a smoother loss landscape compared to FedAvg in the early stages. This smoothness not only enhances the model's generalization but also reduces the risk of getting trapped in local optima, facilitating the discovery of a better global solution.

\begin{figure}[H]
  \centering
  \vspace{-0.3cm}
  \includegraphics[width=0.95\linewidth]{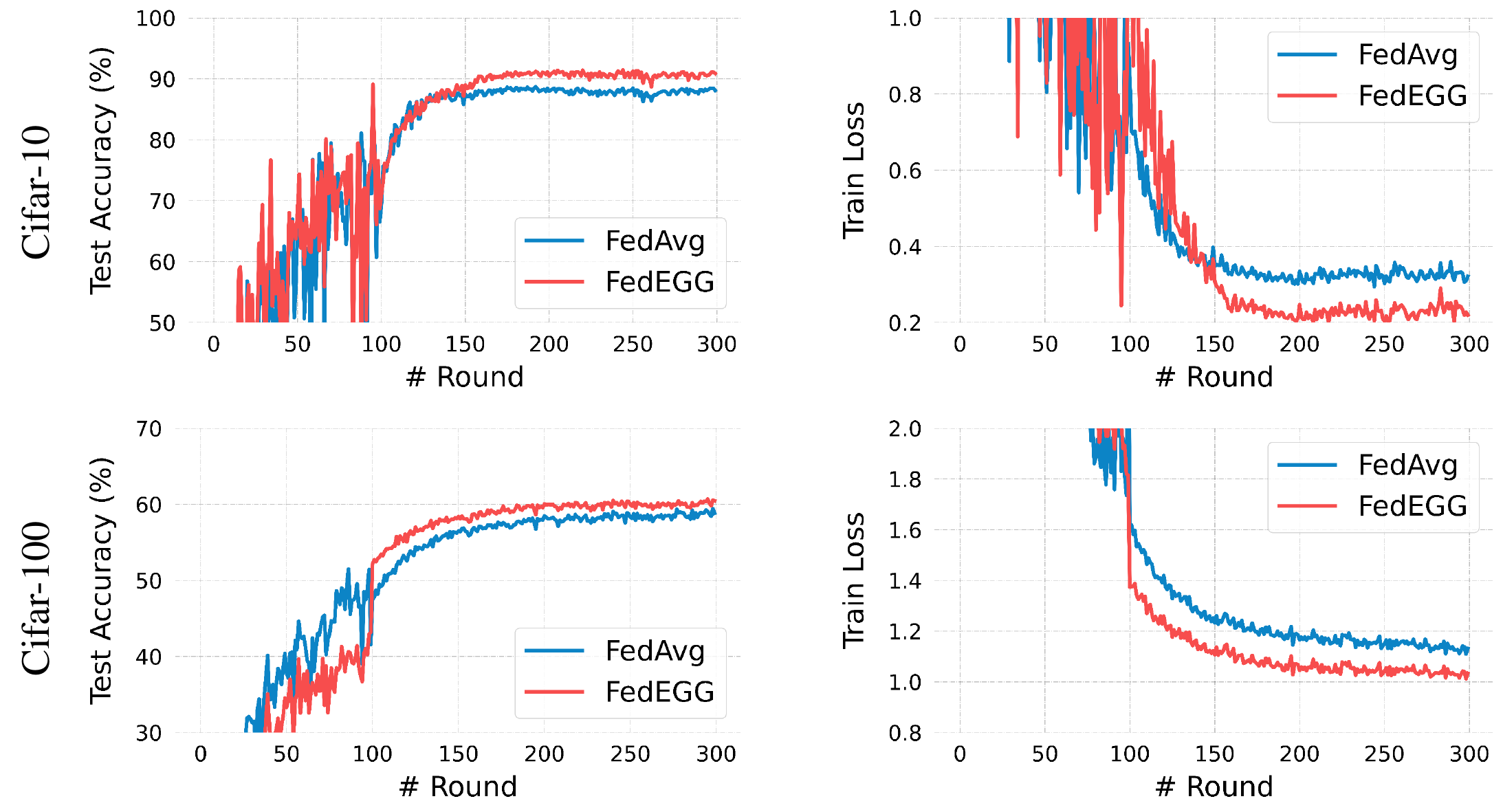}
  \vspace{-0.3cm}
  \caption{ Training losses and test accuracy of FedAvg and FedEGG on CIFAR-10 and CIFAR-100 under the non-IID setting, with synthetic data generated by GPT-4 used as the guiding task for FedEGG.}
  \label{Fig:Synthetic}
  \vspace{-0.3cm}
\end{figure}

\begin{figure*}[!ht]
  \centering
  \includegraphics[width=\linewidth]{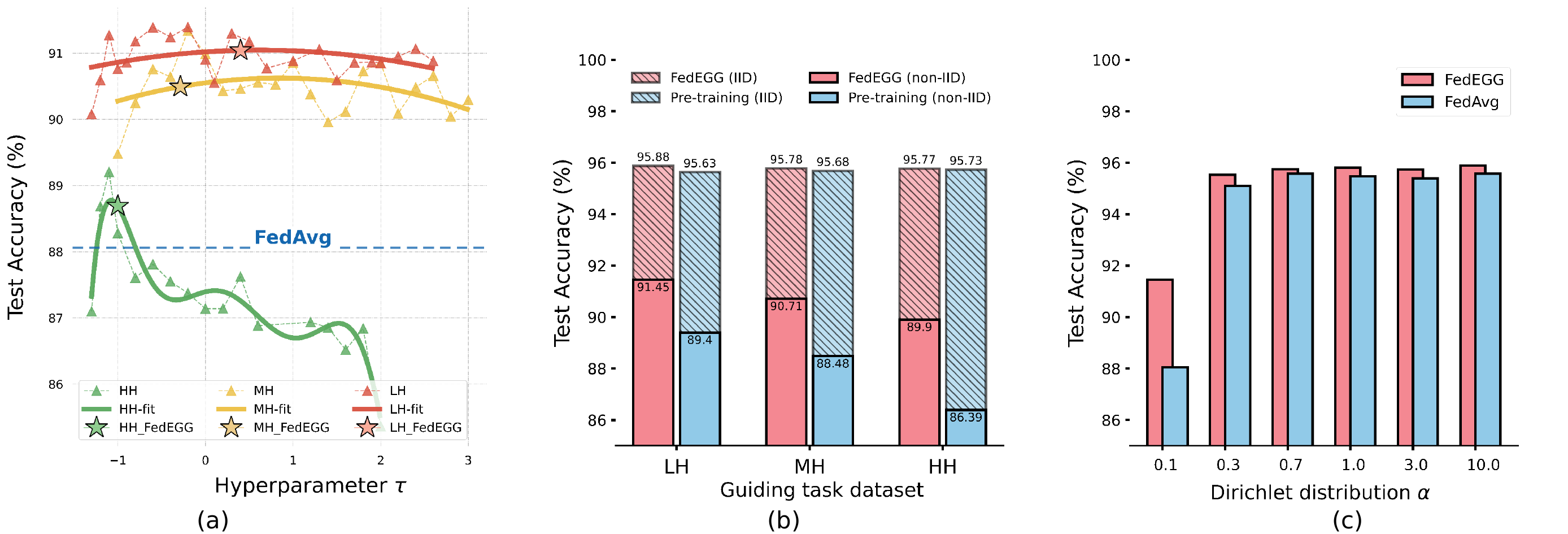}
  \vspace{-0.5cm}
  \caption{\textbf{(a)}: Test accuracy (\%) of FedEGG under different $\tau$ with LH, MH, and HH guiding tasks on CIFAR-10 (non-IID). Triangles denote test accuracy, curves show approximated trends, and asterisks mark $\tau$ calculated in FedEGG. \textbf{(b)}: Test accuracy of FedEGG (red bars) and pre-training (blue bars) on CIFAR-10 under IID and non-IID settings. \textbf{(c)}: Test accuracy of FedEGG and FedAvg under varying levels of FL data heterogeneity $\alpha$ on CIFAR-10.}
  \label{Fig:all}
\end{figure*}

\noindent{\textbf{Observation Regarding Pre-trained Weights. }}\;
One interesting observation is that, when pre-trained weights are used, the classic FedAvg algorithm often outperforms other baselines, particularly on CIFAR-10 under the non-IID setting. Methods like SCAFFOLD and FedProx, which aim to reduce variance among client models by imposing constraints on client learning, are effective when training from scratch but perform poorly with pre-trained weights. We hypothesize that these constraints cause pre-trained models to become trapped in local optima; the stronger the constraint, the more limited the exploration space for clients, resulting in degraded performance. This explains why SCAFFOLD, FedProx, and FedNova underperform compared to FedAvg, and why FedDyn and FedDC lag even further behind. In FL with pre-trained models, the key challenge lies in helping clients escape local optima and adapt quickly to new datasets. Our FedEGG addresses this challenge effectively, demonstrating superior performance in this context.

\noindent{\textbf{Generating Synthetic Guiding Task Data Using MLLM. }}\;
In this experiment, we use text class labels provided by clients and employ GPT-4 to generate synthetic data for the guiding task. To increase the diversity of the generated data, we have implemented a two-stage prompt mechanism. In the first stage, \textit{Prompt 1} directs GPT-4 to create a \textit{Prompt 2} based on a given target [class]. \textit{Prompt 2} is then used for data generation. This approach ensures that the generated data aligns with the target class while maintaining high diversity. Our goal is to preserve class consistency while generating diverse data across different scenes, styles, and variations, thereby enriching the dataset. We applied this approach to generate guiding task data for CIFAR-10 and CIFAR-100, covering 10 and 30 classes respectively, with 32 samples per class. We provide the prompts used and generated image examples in Appendix \ref{sec:D}.
Figure \ref{Fig:Synthetic} presents the test accuracy and training loss of FedEGG on CIFAR-10 and CIFAR-100 under non-iid settings using the synthetic guiding task data. Results show that synthetic data can also accelerates model convergence and improves test accuracy, with test accuracy on CIFAR-10 and CIFAR-100 improving by \textbf{2.76\%} and \textbf{1.45\%}, respectively. These findings validate the feasibility of using generated data for FedEGG.

\subsection{Ablation Studies} \label{sec:Ablation}
\noindent{\textbf{Impact of $\tau $. }}\;
Figure \ref{Fig:all} (a) illustrates the relationship between \(\tau\) and the model's test accuracy on CIFAR-10 under the non-IID setting for different guiding task levels (LH, MH, and HH). Several key trends emerge:  First, as \(\tau\) increases, the model's accuracy initially improves before declining, highlighting the need for an optimal \(\tau\) to avoid  ``underguiding" or ``overguiding". 
Second, the optimal \(\tau\) decreases as task heterogeneity increases (from LH to HH), aligning with our theoretical analysis. This suggests that higher task heterogeneity requires reduced guiding strength to maintain model adaptability and robustness.  
Finally, the asterisks in Figure \ref{Fig:all} (a) represent \(\tau\) values computed using Eq. \eqref{eq-tau} in FedEGG. These values closely match the observed optimal \(\tau\), validating our method for estimating \(\tau\) based on the cosine similarity between guiding and FL task data characteristics. This confirms both our theoretical analysis and the practical effectiveness of our approach.

\noindent{\textbf{Offline vs. Online Guidance. }}\;
FedEGG operates as an online guiding method, where the guiding task is optimized simultaneously with the FL task. To compare, here we test an offline version by pre-training the global model on the guiding task. As shown in Figure \ref{Fig:all} (b), the online FedEGG significantly outperforms the offline approach, particularly under the non-IID setting. Specifically, under the non-IID (Dir(0.1)) setting, FedEGG surpasses offline FedEGG by \textbf{2.05\%}, \textbf{2.23\%}, and \textbf{3.51\%} for guiding task levels LH, MH, and HH, respectively. This highlights the importance of dual optimization of the guiding and FL tasks for improved convergence.

\noindent{\textbf{Effectiveness Across Varying Non-IID Levels. }}\;  
We compare FedEGG (with the LH guiding task) and FedAvg across varying levels of FL data heterogeneity, represented by \(\alpha \in [0.1, 0.3, 0.7, 1, 3, 10]\). As shown in Figure \ref{Fig:all} (c), FedEGG consistently outperforms FedAvg, with its advantage being most pronounced at higher levels of data heterogeneity (e.g., \(\alpha = 0.1\)). This highlights the importance of explicit global guidance in highly heterogeneous data scenarios.

\noindent{\textbf{Impact of Client Participation Rate. }}\;  
Here, we investigate the impact of client participation rate (\(\frac{m}{N} \in [0.1, 0.3, 0.5, 0.7, 1.0]\)) on FedEGG's performance. Table \ref{tab:participation rate} compares FedAvg and FedEGG, revealing that FedEGG generally outperforms FedAvg, especially as task heterogeneity decreases. Lower client participation rates further amplify FedEGG's advantages; for instance, at the lowest participation rate of 0.1 under non-IID conditions, FedEGG (LH) and FedEGG (MH) achieve \textbf{8.22\%} and \textbf{8.53\%} higher test accuracies, respectively, compared to FedAvg. However, at moderate participation rates (0.3 and 0.5), FedEGG (HH) performs slightly worse than FedAvg, indicating areas for improvement. Future work will focus on optimizing FedEGG for these scenarios.

\begin{table}[!htb]
  \vspace{-0.2cm}
  \caption{Average test accuracies (\%) with varying sample rates on CIFAR-10 under non-IID and IID settings. The +/- symbols indicate the increase or decrease in performance of FedEGG when compared with FedAvg.}
  \label{tab:participation rate}
  \resizebox{1\linewidth}{!}{ 
   \centering
  \begin{tabular}{c | c  |c | c | c| c  |  c | c | c |c | c| c}
    \toprule
   &  \multirow{2}{*}{\diagbox{\textbf{Method}}{\textbf{Sample}}}  &  \multirow{2}{*}{\textbf{0.1}}   &  \multirow{2}{*}{\textbf{0.3}} & \multirow{2}{*}{\textbf{0.5}} & \multirow{2}{*}{\textbf{0.7}}  & \multirow{2}{*}{\textbf{1}}  \\ 
   &     &  &    &   &    &  \\
    \midrule
   \multirow{4}{*}{\rotatebox{90}{\textbf{Non-IID}}}  &  \textbf{FedAvg}    &    70.7  &     88.32       &    88.69      &   88.38      & 90.2  \\
   &  \textbf{FedEGG (LH)}    &    78.916 (+8.22)  &      89.71 (+1.39)  &    91.47 (+2.78)  &      91.58 (+3.20)   &   90.4 (+0.2)    \\
   & \textbf{FedEGG (MH)}    &    79.24 (+8.53)  &     88.92 (+0.60)   &       88.78 (+0.09)  &       91.09 (+2.71)   &     90.9 (+0.7)   \\
   & \textbf{FedEGG (HH)}    &    71.94 (+1.24)  &     87.38 (-0.93)   &      88.32 (-0.38)  &       90.01 (+1.63)       &  90.37 (+0.17)   \\
    \midrule
   \multirow{4}{*}{\rotatebox{90}{\textbf{IID}}} & \textbf{FedAvg}     & 93.96  &    94.69     &    94.78      &   94.88   &  94.82    \\
   
   & \textbf{FedEGG (LH)}    &  94.58 (+0.62)  &   95.25 (+0.56)    &    95.23 (+0.45) &   95.29 (+0.41)  &  95.38 (+0.56)     \\
   
   & \textbf{FedEGG (MH)}    &     94.26 (+0.3) &   95.18 (+0.49)    &    95.13 (+0.349)  &   95.26 (+0.38)  &  95.23 (+0.41)    \\
   
   & \textbf{FedEGG (HH)}    &   94.336 (+0.376)   &  94.78 (+0.09)   &    94.81 (+0.03)  &   94.94 (+0.159)   & 94.92 (+0.1)    \\
    \bottomrule    
  \end{tabular}
  }
\end{table}

\section{Conclusion}
In this paper, we proposed a Federated Learning (FL) algorithm named \textbf{FedEGG} that introduces an explicit global guiding task to enhance the convergence and performance of FL under non-IID settings. By constructing a well-defined, easy-to-converge server-side guiding task based on public datasets and Large Language Models (LLMs), FedEGG effectively combines the strengths of federated and centralized learning. Our theoretical analysis provided insights into the impact of task heterogeneity and guiding strength, deriving an upper bound for optimal guidance. Empirically, FedEGG demonstrated superior performance over state-of-the-art FL methods and further improved their performance when combined. These results highlight the potential of explicit global guidance in addressing FL convergence challenges, particularly in real-world scenarios with heterogeneous data. Future work will focus on optimizing FedEGG for varying network conditions and exploring its applicability to broader domains.

\bibliography{example_paper}
\bibliographystyle{icml2025}

\newpage
\appendix
\onecolumn

\section{Proof of Theorem 1. }\label{sec:A}
\begin{proof}

Notice that $\mathbf{W}_{t+1} = \mathbf{W}_t - \eta_t\mathbf{g}_t -\gamma_t \boldsymbol{q}_t $
\begin{flalign*}
    & \ \ \ \  \|\mathbf{W}_{t+1} - \mathbf{W}^* \|^2 \\
     & = \| \mathbf{W}_t - \eta_t\mathbf{g}_t -\gamma_t \boldsymbol{q}_t - \mathbf{W}^* \|^2 \\
     &= \| \mathbf{W}_t - \eta_t\overline{\mathbf{g}}_t -\gamma_t\overline{\boldsymbol{q}}_t - \mathbf{W}^* + \eta_t\overline{\mathbf{g}}_t - \eta_t\mathbf{g}_t +\gamma_t\overline{\boldsymbol{q}}_t -\gamma_t \boldsymbol{q}_t \|^2 \\
     & = \underbrace{\| \mathbf{W}_t - \eta_t\overline{\mathbf{g}}_t -\gamma_t\overline{\boldsymbol{q}}_t - \mathbf{W}^* \|^2}_{A1}  + \underbrace{ \| \eta_t\overline{\mathbf{g}}_t - \eta_t\mathbf{g}_t +\gamma_t\overline{\boldsymbol{q}}_t -\gamma_t \boldsymbol{q}_t \|^2}_{A2} + \underbrace{2 \langle \mathbf{W}_t - \eta_t\overline{\mathbf{g}}_t -\gamma_t\overline{\boldsymbol{q}}_t - \mathbf{W}^* ,  \eta_t\overline{\mathbf{g}}_t - \eta_t\mathbf{g}_t +\gamma_t\overline{\boldsymbol{q}}_t -\gamma_t \boldsymbol{q}_t \rangle}_{A3}
\end{flalign*}

Note that $\mathbb{E}[A3]=0$. We next focus on bounding $A1$ and $A2$.
\begin{flalign*}
    A1 & = \| \mathbf{W}_t - \eta_t\overline{\mathbf{g}}_t - \gamma_t\overline{\boldsymbol{q}}_t - \mathbf{W}^* \|^2   \\
       & = \| \mathbf{W}_t - \eta_t\overline{\mathbf{g}}_t - \mathbf{W}^* \|^2 + \gamma_t^2 \| \overline{\boldsymbol{q}}_t \|^2 - 2 \gamma_t \langle \mathbf{W}_t - \eta_t\overline{\mathbf{g}}_t - \mathbf{W}^*, \overline{\boldsymbol{q}}_t \rangle
\end{flalign*}
\vspace{-0.2cm}
\begin{align*}
    A2 & = \| \eta_t\overline{\mathbf{g}}_t - \eta_t\mathbf{g}_t +\gamma_t\overline{\boldsymbol{q}}_t -\gamma_t \boldsymbol{q}_t \|^2  \\
    & = \eta_t^2 \| \overline{\mathbf{g}}_t - \mathbf{g}_t  \|^2 + \gamma_t^2 \| \overline{\boldsymbol{q}}_t-  \boldsymbol{q}_t  \|^2 + 2\gamma_t \eta_t \langle \overline{\mathbf{g}}_t - \mathbf{g}_t, \overline{\boldsymbol{q}}_t-  \boldsymbol{q}_t \rangle \\
    & = \eta_t^2 \| \overline{\mathbf{g}}_t - \mathbf{g}_t  \|^2 + \gamma_t^2 \| \overline{\boldsymbol{q}}_t-  \boldsymbol{q}_t  \|^2
\end{align*}

Then
\begin{flalign*}
   & \ \ \ \ \ \|\mathbf{W}_{t+1} - \mathbf{W}^* \|^2  \\
    & =\| \mathbf{W}_t - \eta_t\overline{\mathbf{g}}_t - \mathbf{W}^* \|^2
       -2 \gamma_t \langle \mathbf{W}_t - \eta_t\overline{\mathbf{g}}_t - \mathbf{W}^* , \overline{\boldsymbol{q}}_t  \rangle + \eta_t^2 \| \overline{\mathbf{g}}_t - \mathbf{g}_t  \|^2 + \gamma_t^2 \| \overline{\boldsymbol{q}}_t-  \boldsymbol{q}_t  \|^2 + +  \gamma_t^2 \| \overline{\boldsymbol{q}}_t\|^2\\
    & = \| \mathbf{W}_t - \eta_t\overline{\mathbf{g}}_t - \mathbf{W}^* \|^2 + \eta_t^2 \| \overline{\mathbf{g}}_t - \mathbf{g}_t  \|^2 + \gamma_t^2 \| \overline{\boldsymbol{q}}_t\|^2  -2 \eta_t \langle  \mathbf{W}_t - \eta_t\overline{\mathbf{g}}_t - \mathbf{W}^*,  \overline{\mathbf{g}}_t - \mathbf{g}_t \rangle  -2 \gamma_t \langle \mathbf{W}_t - \eta_t\overline{\mathbf{g}}_t - \mathbf{W}^* , \overline{\boldsymbol{q}}_t  \rangle \\
      & \ \ \ \ + 2 \eta_t \langle  \mathbf{W}_t - \eta_t\overline{\mathbf{g}}_t - \mathbf{W}^*,  \overline{\mathbf{g}}_t - \mathbf{g}_t \rangle + \gamma_t^2 \| \overline{\boldsymbol{q}}_t-  \boldsymbol{q}_t  \|^2 \\
   & = \| \mathbf{W}_t - \eta_t\overline{\mathbf{g}}_t - \mathbf{W}^* + \eta_t \overline{\mathbf{g}}_t - \eta_t\mathbf{g}_t \|^2  -2 \gamma_t \langle \mathbf{W}_t - \eta_t\overline{\mathbf{g}}_t - \mathbf{W}^* , \overline{\boldsymbol{q}}_t  \rangle   -2 \eta_t \langle  \mathbf{W}_t - \eta_t\overline{\mathbf{g}}_t - \mathbf{W}^*,  \overline{\mathbf{g}}_t - \mathbf{g}_t \rangle  + \gamma_t^2 \| \overline{\boldsymbol{q}}_t\|^2 \\
    & \ \ \ \ + \gamma_t^2 \| \overline{\boldsymbol{q}}_t-  \boldsymbol{q}_t  \|^2 \\
    & =  \| \mathbf{W}_t - \eta_t\overline{\mathbf{g}}_t - \mathbf{W}^* + \eta_t \overline{\mathbf{g}}_t - \eta_t\mathbf{g}_t \|^2  -2 \gamma_t \langle \mathbf{W}_t - \eta_t\overline{\mathbf{g}}_t - \mathbf{W}^* , \overline{\boldsymbol{q}}_t  \rangle + \gamma_t^2 \| \overline{\boldsymbol{q}}_t\|^2 + \gamma_t^2 \| \overline{\boldsymbol{q}}_t-  \boldsymbol{q}_t  \|^2 \\  
     & = \underbrace{ \| \mathbf{W}_t - \mathbf{W}^* -\eta_t\mathbf{g}_t \|^2 }_{FedAvg} \underbrace{-2 \gamma_t \langle \mathbf{W}_t - \eta_t\overline{\mathbf{g}}_t - \mathbf{W}^* , \overline{\boldsymbol{q}}_t  \rangle }_{B1}  + \gamma_t^2 \| \overline{\boldsymbol{q}}_t\|^2 + \gamma_t^2 \| \overline{\boldsymbol{q}}_t-  \boldsymbol{q}_t  \|^2
\end{flalign*}

where $\mathbb{E}(\langle  \mathbf{W}_t - \eta_t\overline{\mathbf{g}}_t - \mathbf{W}^*,  \overline{\mathbf{g}}_t - \mathbf{g}_t \rangle) =0$.

Next, we focus on bounding $B1$,
\begin{flalign*}
   B1 &= -2 \gamma_t \langle \mathbf{W}_t - \eta_t\overline{\mathbf{g}}_t - \mathbf{W}^* , \overline{\boldsymbol{q}}_t  \rangle \\
   & = -2 \gamma_t \langle \overline{\mathbf{W}}_t - \mathbf{W}^* , \overline{\boldsymbol{q}}_t \rangle \\
   & = \underbrace{-2 \gamma_t \langle \overline{\mathbf{W}}_t -\mathbf{W}_g^* , \nabla F(\overline{\mathbf{W}}_t)\rangle}_{C1}  \underbrace{-2\gamma_t \langle \mathbf{W}_g^* - \mathbf{W}^* , \nabla F(\overline{\mathbf{W}}_t)\rangle}_{C2}
\end{flalign*}

\vspace{-10pt}
\begin{flalign*}
    C1 & = -2 \gamma_t \langle \overline{\mathbf{W}}_t -\mathbf{W}_g^* , \nabla F(\overline{\mathbf{W}}_t) \rangle \leq 2\gamma_t (F^* - F(\overline{\mathbf{W}}_t)) - \frac{\gamma_t}{L}\|  \overline{\boldsymbol{q}}_t  \|^2
\end{flalign*}

\vspace{-10pt}

\begin{flalign*}
    C2 & = -2\gamma_t \langle \mathbf{W}_g^* - \mathbf{W}^* , \nabla F(\overline{\mathbf{W}}_t)\rangle    \\
    & \leq \| \mathbf{W}_g^* - \mathbf{W}^*  \|^2  + \gamma_t^2 \|  \overline{\boldsymbol{q}}_t  \|^2  \\
    & \leq \frac{2}{\mu} \big(F(\mathbf{W}^* )  -  F(\mathbf{W}_g^* ) \big)  + \gamma_t^2 \|  \overline{\boldsymbol{q}}_t\|^2 \\
    & = \frac{2}{\mu} \big(f^*  -  F^* \big)  + \gamma_t^2 \|  \overline{\boldsymbol{q}}_t\|^2  \\
    & = \frac{2}{\mu}\Gamma_g + \gamma_t^2 \|  \overline{\boldsymbol{q}}_t\|^2,
\end{flalign*}
where the first inequality is a result from $AM-GM$ inequality, the second inequality is from $\mu-strong$ assumption, and we use the notation $\Gamma_g= f^*-F^*$.

Then, we have
\begin{flalign*}
    B1 \leq (\gamma_t^2 - \frac{\gamma_t}{L})\| \overline{\boldsymbol{q}}_t\|^2 + \frac{2}{\mu}\Gamma_g  +2\gamma_t (F^* - F(\overline{\mathbf{W}}_t))
\end{flalign*}

So,
\begin{flalign*}
     \mathbb{E} \|\mathbf{W}_{t+1} - \mathbf{W}^* \|^2 = \underbrace{ \mathbb{E} \| \mathbf{W}_t - \mathbf{W}^* -\eta_t\mathbf{g}_t\|^2 }_{FedAvg}  \underbrace{+ \frac{2}{\mu}\Gamma_g + \gamma_t^2 \mathbb{E} \| \overline{\boldsymbol{q}}_t-  \boldsymbol{q}_t  \|^2 - 2\gamma_t ( F(\overline{\mathbf{W}}_t)- F^* )}_{D1}  \underbrace{- ( \frac{\gamma_t}{L} -2\gamma_t^2 )\mathbb{E} \| \overline{\boldsymbol{q}}_t\|^2}_{D2} \\
\end{flalign*}
\vspace{-0.6cm}
\begin{flalign*}
  D1+ D2  & = \gamma_t^2 \mathbb{E} \big[ \| \overline{\boldsymbol{q}}_t-  \boldsymbol{q}_t  \|^2 \big] + \frac{2}{\mu}\Gamma_g  -2\gamma_t ( F(\overline{\mathbf{W}}_t)- F^*)  -  ( \frac{\gamma_t}{L} - 2\gamma_t^2 ) \mathbb{E}\big[\| \overline{\boldsymbol{q}}_t\|^2\big] \\
    & = \gamma_t^2\sigma_g^2 - (4\gamma_t - 4L\gamma_t^2)(F(\overline{\mathbf{W}}_t))- F^*) + \frac{2}{\mu} \Gamma_g \\
    & =  \gamma_t^2\sigma_g^2  + (\frac{2}{\mu}  + 4L\gamma_t^2 - 4\gamma_t ) \Gamma_g  - (4\gamma_t - 4L\gamma_t^2)  (F(\overline{\mathbf{W}}_t))- f^*)
\end{flalign*}

The result is
\begin{flalign*}
 \mathbb{E} \|\mathbf{W}_{t+1} - \mathbf{W}^* \|^2 
    & = \underbrace{ \mathbb{E} \| \mathbf{W}_t- \mathbf{W}^* -\eta_t\mathbf{g}_t\|^2 }_{FedAvg}   +  \gamma_t^2\sigma_g^2  + (\frac{2}{\mu}  + 4L\gamma_t^2 - 4\gamma_t ) \Gamma_g + (4\gamma_t - 4L\gamma_t^2) \Pi,
\end{flalign*}
where $\Pi =  f^* -F(\overline{\mathbf{W}}_t) $.

\end{proof}

\section{Guiding Task Construction for FedEGG}\label{sec:B}

\begin{table}[!ht]
  \caption{The top 4 most relevant ImageNet labels found by GPT-4 for each CIFAR-10 class.}
  \label{tab:1}
  \small
  \centering
  \begin{tabular}{cl}
    \toprule
    CIFAR-10    &   ImageNet-1k  \\
    \midrule
    airplane     &   \{"n04552348", "n02690373", "n02687172", "n04552348" \}    \\
    automobile    &   \{"n02814533", "n03100240", "n03594945", "n04467665" \} \\
    bird    &  \{"n01796340", "n01797886", "n01608432", "n01795545"\}   \\
    cat   &  \{"n02124075", "n02123597", "n02123045", "n02123394"\}   \\
    deer   &  \{"n02437616", "n02412080", "n02415577", "n02396427" \}  \\
    dog   & \{"n02085782", "n02085620", "n02085620", "n02085936" \}   \\
    frog   &  \{"n01644900", "n01644373", "n01641577", "n01910747" \}   \\
    horse   &  \{"n02389026", "n02391049", "n02412080", "n02504458" \}   \\
    ship   &  \{"n03673027", "n02981792", "n03095699", "n03673027" \}   \\
    truck   &  \{"n03417042", "n04467665", "n03777568", "n03770679” \}   \\
    \bottomrule
  \end{tabular}
\end{table}
We assume the server has access to meta-information about the FL task, such as the class names of the classification task. If class names are unavailable, the server can request clients to upload their associated concepts. Using this meta-information, the server can identify the most similar concepts in a public dataset (e.g., ImageNet-1k) with the assistance of GPT-4. In scenarios where no closely related public datasets are available, the server can utilize a diffusion model to generate images representing similar concepts. However, in this work, we focus exclusively on the case where a large-scale public dataset is available to construct the guiding task. Below, we detail the selection process for the CIFAR-10 guiding tasks, which can be similarly applied to the CIFAR-100 dataset.

\noindent \textbf{Constructing guiding task for CIFAR-10} \;
Here, we present the prompt and GPT-4 response used to identify the most similar classes in the ImageNet-1k dataset for the CIFAR-10 FL task.

\paragraph{Prompt: } ``For each concept in the following `Client Label List', please select the top four most relevant labels from the above `Reference Label List'. Please sort the selected labels for each concept and return them as a dictionary with the concept as the key and the top 4 most relevant labels as a list of values. You should return 40 labels in total. Client Label List: \{1: 'airplane', 2: 'automobile', 3: 'bird', 4: 'cat', 5: 'deer', 6: 'dog', 7: 'frog', 8: 'horse', 9: 'ship', 10: 'truck'\}."

\paragraph{GPT-4 Answer: } \{'airplane': lable1, label2, label3, label4. \}, ..., \{'truck': lable1, label2, label3, label4. \}

The selected ImageNet classes are listed in Table \ref{tab:1}. We then select the top 20 classes to construct the Low Heterogeneity (LH) guiding task. For the Medium Heterogeneity (MH) guiding task, we replace the top 10-20 classes with 10 randomly selected ImageNet classes. For the High Heterogeneity (HH) guiding task, we randomly select 20 classes from ImageNet-1k.

\section{Training Losses on CIFAR-10 and CIFAR-100}\label{sec:C}
\begin{figure*}[!ht]
      \centering
  \includegraphics[width=\linewidth]{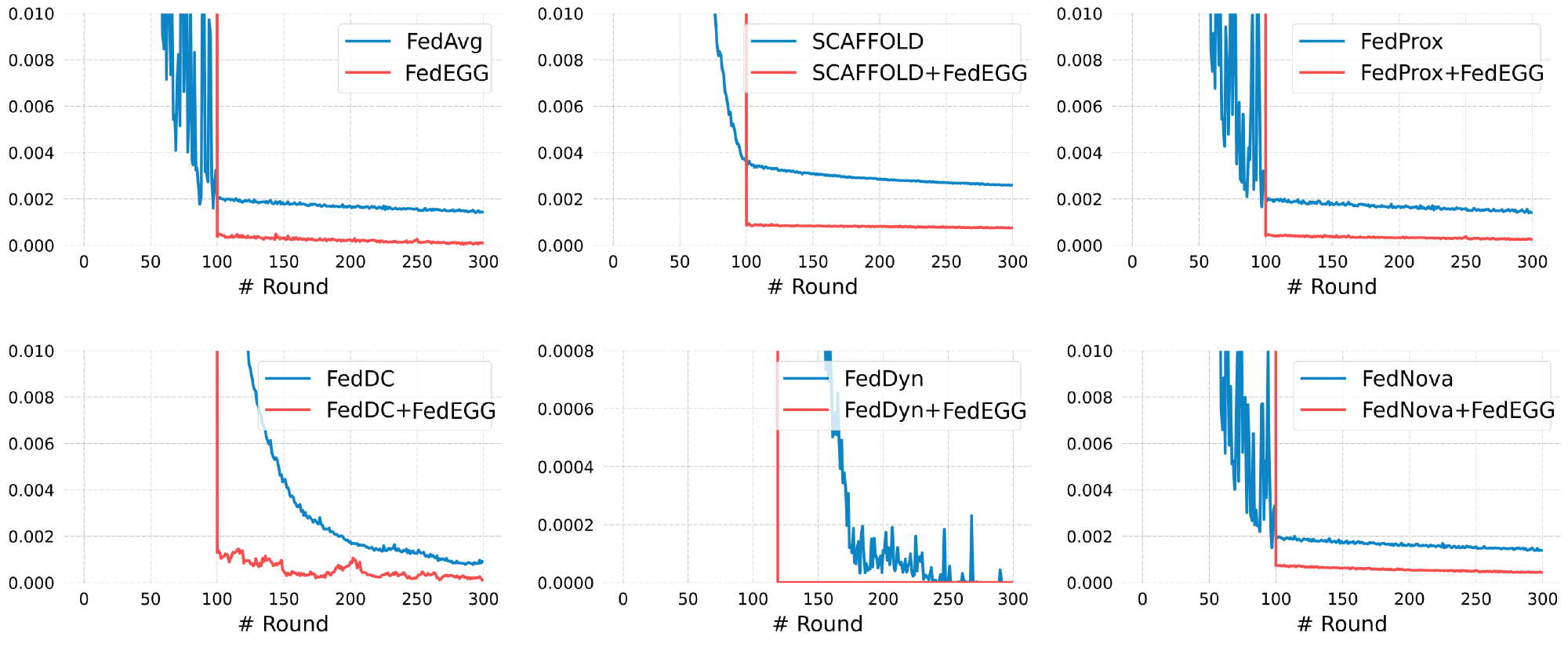}
  \caption{The training losses of different FL methods across 300 rounds on CIFAR-10 under the IID setting.}
  \label{Fig:cifar10_iid}
\end{figure*}

Figure \ref{Fig:cifar10_iid} illustrates the training loss of our FedEGG and other baseline methods on CIFAR-10 under the IID setting. It is evident that our method enhances convergence even in the IID setting, with this advantage being universally consistent across different FL methods. Surprisingly, the convergence improvement is more pronounced under the IID setting compared to the non-IID setting (shown in Figure 1 of the main text). This is because optimization on IID data is inherently more stable than on non-IID data, allowing the global guidance to work effectively and consistently across all communication rounds. In contrast, the non-IID setting is less stable, requiring careful selection and strength control of the guiding task.

\begin{figure*}[!ht]
      \centering
  \includegraphics[width=\linewidth]{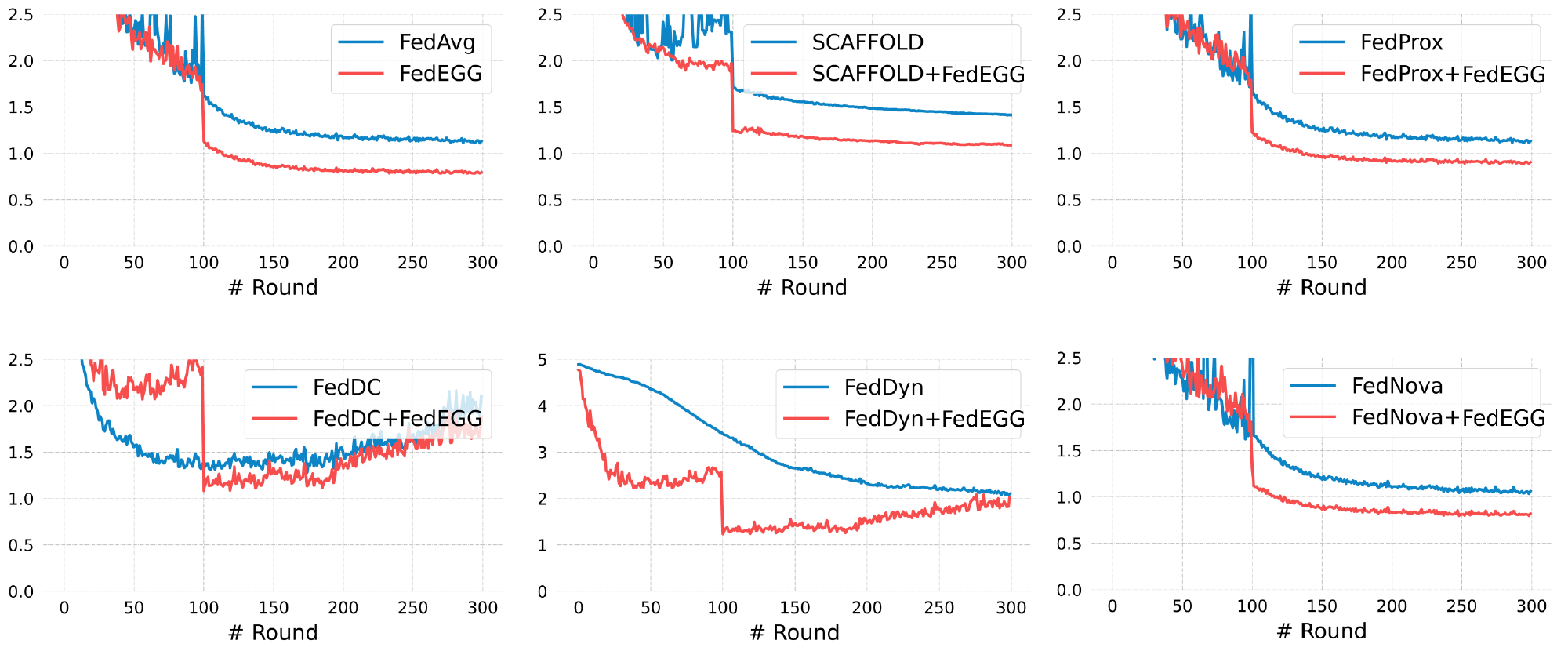}
  \caption{ The training losses of different FL methods across 300 rounds on CIFAR-100 under the non-IID setting.}
  \label{Fig:cifar100_noniid}
\end{figure*}

\begin{figure*}[!ht]
      \centering
  \includegraphics[width=\linewidth]{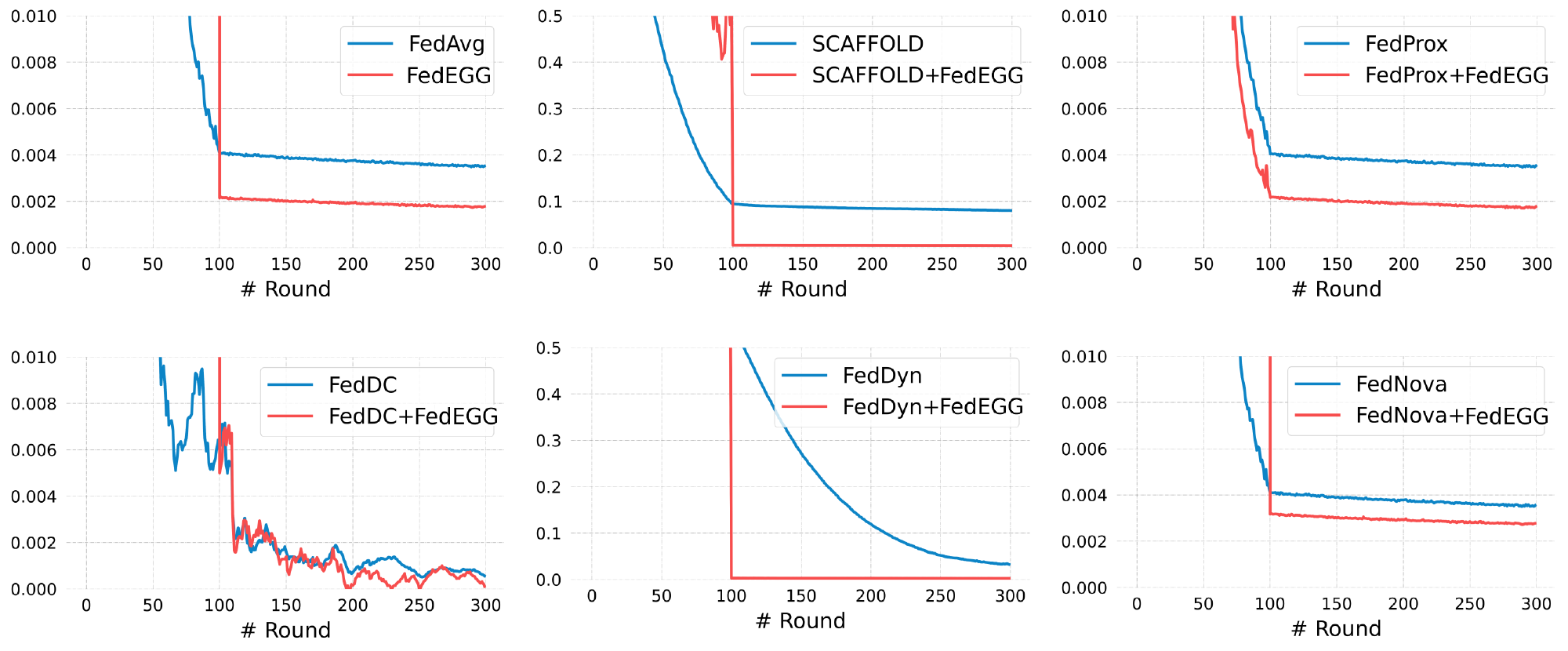}
  \caption{ The training losses of different FL methods across 300 rounds on CIFAR-100 under the IID setting.}
  \label{Fig:cifar100_iid}
\end{figure*}

Figures \ref{Fig:cifar100_noniid} and \ref{Fig:cifar100_iid} further illustrate the convergence curves of our FedEGG on CIFAR-100 under non-IID and IID settings, respectively. 

\section{Generation Details of Synthetic Guiding Task Data}\label{sec:D}
This experiment aims to generate synthetic data using client-side text labels through GPT-4, addressing the limitations of existing publicly available datasets in supporting guiding tasks. To overcome this challenge, we propose a two-stage prompt mechanism.

First, we create \textit{prompt 1}, which includes specific text category information to guide GPT-4 in generating \textit{prompt 2}. \textit{Prompt 2} must ensure both category accuracy and data diversity. Finally, prompt2 is input into GPT-4, which generates the corresponding synthetic data.

The following provides a detailed example of the generation process using the text category "ship":

\paragraph{Prompt 1:} Create a prompt to guide the LLM in generating 64x64 pixel images of the category "ship". The images should accurately reflect the category and feature diverse content, including various species, scenes, and angles, all presented in a real-life photography style.

\paragraph{GPT-4 Answer:} 
Generate a series of 64x64 pixel images of ships, showcasing a variety of ship types, scenes, and angles, all in a real-life photography style. Specific requirements are as follows:

(1) Ship Types:
\begin{itemize}
    \item Modern ships: cargo ships, cruise ships, warships, oil tankers, luxury yachts, fishing boats, ferries, submarines, speedboats.
    \item Historical ships: ancient warships, sailing ships, pirate ships, Victorian-era ships, wooden boats, steamships, old-style sailing vessels.
    \item  Ships from different regions: ships in the Arctic with snow, tropical fishing boats, riverboats, inland lake ships.
    \item  Specialized ships: rescue ships, fireboats, hospital ships, research vessels, military landing crafts, icebreakers, submarines.
    \item Ships with various functions: small cargo ships, large passenger ships, transport ships, luxury liners.
\end{itemize}

(2) Different Angles and Perspectives:
\begin{itemize}
    \item Top view, side view, front view, rear view, oblique view of ships.
    \item Ships in the open ocean, ships docked at a dock, ships in a harbor, ships near a deserted island.
    \item Close-up shots showing details of the ship, such as the hull, mast, anchor, and bow.
    \item Dynamic scenes of ships in motion: sailing through waves, heading into the sunset, navigating rough seas, passing through snow-capped mountains.
\end{itemize}

(3) Various Environments and Weather:
\begin{itemize}
    \item Different oceanic environments: turbulent seas, calm lakes, icy waters, sailing through a storm.
    \item Various weather conditions: morning fog, evening twilight, sunny days, rainy weather, cloudy skies, and thunderstorms.
    \item Different seasons: winter ice-bound ships, summer sailing, autumn harbor scenes.
    \item Natural elements: seagulls flying, waves crashing, ships meeting the horizon.
\end{itemize}

(4) Natural Lighting and Realistic Textures:
\begin{itemize}
    \item Use natural lighting effects, such as sunlight reflections, night-time lights, or the golden glow of sunset.
    \item Detailed textures of the ships: the shine of metal, the grains of wood, the wrinkles of sails, reflections on the waves, rust on old ships.
    \item Add relevant background elements around the ships, such as the ocean, islands, distant mountains, or fishing villages.
\end{itemize}

The images should maintain high quality, clearly depicting the diversity of ships and scenes while preserving a realistic photographic style that reflects the grandeur of ship travel.

\begin{figure*}[!ht]
    \centering
    \includegraphics[width=\linewidth]{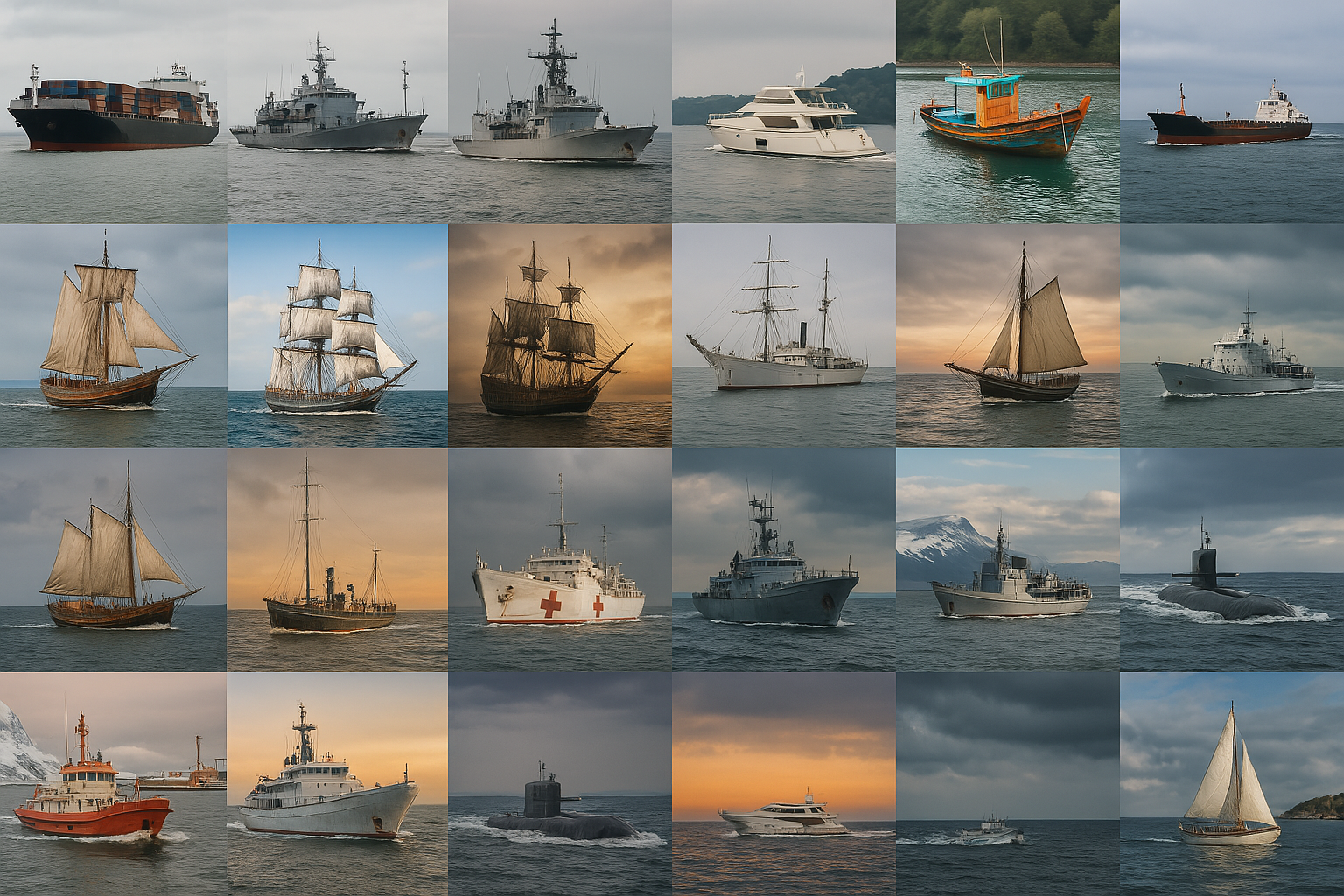}
    \caption{Synthetic data for the 'ship' category generated by GPT-4}
    \label{fig: synthetic}
\end{figure*}

Next, we use \textit{GPT-4 Answer} as \textit{prompt 2} to instruct GPT-4 to generate synthetic data, as shown in Figure \ref{fig: synthetic}.

\end{document}